\documentclass[runningheads,a4paper]{llncs}
\usepackage{amsmath}
\usepackage{amsfonts}
\usepackage{amssymb}
\usepackage{graphicx}
\usepackage{enumerate}
\usepackage[capitalise]{cleveref}
\crefname{section}{Sect.}{Sect.}
\crefname{equation}{}{}
\usepackage{bm}
\usepackage{url}
\usepackage{tikz}
\usetikzlibrary{positioning}
\usetikzlibrary{shapes.geometric}
\usepackage{circuitikz}
\usepackage{algorithm}
\usepackage{algpseudocode}

\algnewcommand\algorithmicforeach{\textbf{for each}}
\algdef{S}[FOR]{ForEach}[1]{\algorithmicforeach\ #1\ \algorithmicdo}

\makeatletter
\AtBeginDocument{%
  \def\doi#1{\url{https://doi.org/#1}}}
\makeatother

\newcommand{\R}[0]{\mathbb{R}}
\newcommand{\C}[0]{\mathbb{C}}
\newcommand{\Q}[0]{\mathbb{Q}}
\newcommand{\N}[0]{\mathbb{N}}
\newcommand{\Z}[0]{\mathbb{Z}}

\newcommand{\cF}[0]{\mathcal{F}}
\newcommand{\cP}[0]{\mathcal{P}}
\newcommand{\cS}[0]{\mathcal{S}}
\newcommand{\cV}[0]{\mathcal{V}}
\newcommand{\Abar}{\bar{A}}
\newcommand{\cbar}[0]{\bar{c}}
\newcommand{\Xbar}{\bar{X}}

\newcommand{\lt}{\mathrm{LT}}
\newcommand{\lm}{\mathrm{LM}}
\newcommand{\lc}{\mathrm{LC}}
\newcommand{\ideal}[1]{{\langle{#1}\rangle}}
\newcommand{\Lplus}[1]{L_{{#1}+}}
\newcommand{\Lminus}[1]{L_{{#1}-}}
\newcommand{\Lbarplus}[1]{\bar{L}_{{#1}+}}
\newcommand{\Lbarminus}[1]{\bar{L}_{{#1}-}}
\newcommand{\Splus}[1]{S_{{#1}+}}
\newcommand{\Sminus}[1]{S_{{#1}-}}

\newcommand{\rmSigma}{\mathrm{\Sigma}}

\newcommand{\disc}[1]{\mathrm{disc}({#1})}
\newcommand{\sdot}{\dot{s}}
\newcommand{\sddot}{\ddot{s}}
\newcommand{\p}{\bm{p}}
\newcommand{\T}{{}^t}
\newcommand{\FAIL}{FAIL}
\newcommand{\CallGenerateRealCGS}{{\textrm{RealCGS}}}
\newcommand{\TRUE}{\textrm{TRUE}}
\newcommand{\FALSE}{\textrm{FALSE}}

\begin{document}
\title{
    Inverse Kinematics and Path Planning of Manipulator Using Real Quantifier Elimination 
    Based on Comprehensive Gr\"obner Systems
}
\titlerunning{Inverse Kinematics and Path Planning of Manipulator Using CGS-QE}
\author{
    Mizuki Yoshizawa\inst{1}\and
    Akira Terui\inst{1}\orcidID{0000-0003-0846-3643} \and
    Masahiko Mikawa\inst{1}\orcidID{0000-0002-2193-3198}
}
\institute{%
University of Tsukuba, Tsukuba, Japan \\
\email{terui@math.tsukuba.ac.jp}\\
\email{mikawa@slis.tsukuba.ac.jp}\\
\url{https://researchmap.jp/aterui}
}

\maketitle

\begin{abstract}
    Methods for inverse kinematics computation and path planning of a three degree-of-freedom (DOF)
    manipulator using the algorithm for quantifier elimination based on Comprehensive 
    Gr\"obner Systems (CGS), called CGS-QE method, are proposed. The first method for solving the 
    inverse kinematics problem employs counting the real roots of a system of polynomial equations 
    to verify the solution's existence.
    In the second method for trajectory planning of the manipulator, the use of 
    CGS guarantees the existence of an inverse kinematics solution. Moreover, it
    makes the algorithm more efficient by preventing repeated computation of Gr\"obner basis. 
    In the third method for path planning of the manipulator, for a path of the motion given as
    a function of a parameter, the CGS-QE method verifies the whole path's feasibility. Computational examples and an experiment are provided to illustrate the effectiveness 
    of the proposed methods.
\end{abstract}

\section{Introduction}

We discuss inverse kinematics computation of a 3-degree-of-freedom (DOF) manipulator 
using computer algebra.
Manipulator is a robot with links and joints that are connected alternatively. 
The end part is called the end-effector.
The inverse kinematics problem is fundamental in motion planning.
In the motion planning of manipulators, 
a mapping from a joint space and the operational space of the end-effector
is considered for solving the forward and inverse kinematics problems.
The forward kinematics problem is solved to find the end-effector's position from
the given configuration of the joints.
On the other hand, the inverse kinematic problem is solved to find
the configuration of the joints if the solution exists.

For solving inverse kinematics problems, computer algebra methods have been 
proposed \cite{fau-mer-rou2006,kal-kal1993,ric-sch-ces2021,uch-mcp2011,uch-mcp2012}. 
Some of these methods are especially for modern manipulators with 
large degrees of freedom \cite{ric-sch-ces2021}, 
which indicates an interest in applying global methods to a real-world problem.
The inverse kinematics problem is expressed as a system of polynomial
equations in which trigonometric functions are replaced with variables, 
and constraints on the trigonometric functions are added as new equations.
Then, the system of equations gets ``triangularized'' by computing a
Gr\"obner basis and approximate solutions are calculated using appropriate solvers.
We have proposed an implementation for inverse kinematics computation of
a 3-DOF manipulator \cite{hor-ter-mik2020}. 
The implementation uses SymPy, a library of computer algebra, on top of Python,
and also uses a computer algebra system Risa/Asir \cite{nor2003}
for Gr\"obner basis computation,
connected with OpenXM infrastructure \cite{mae-nor-oha-tak-tam2001}.

An advantage of using Gr\"obner basis computation for solving 
inverse kinematics problems is that the global solution can be obtained.
The global solution helps to characterize the  
robot's motion, such as kinematic singularities.
On the other hand, Gr\"obner basis computation is relatively costly.
Thus, repeating Gr\"obner basis computation every time the position 
of the end-effector changes leads to an increase in computational cost.
Furthermore, in inverse kinematics computation with a global method,
it is necessary to determine if moving the 
end-effector to a given destination is feasible.
Usually, numerical methods are used to compute an approximate solution 
of the system of polynomial equations, but this is only 
an approximation and another computation is required 
to verify the existence of the solution to the inverse kinematics problem.
In fact, our previous implementation above
has the problem of calculating approximate solutions without 
verifying the existence of the real solution to the inverse kinematics problem.

We have focused on Comprehensive Gr\"obner Systems (CGS).
CGS is a theory and method for computing Gr\"obner bases 
for ideals of the polynomial ring, where generators of the ideal have
parameters in their coefficients.
Gr\"obner basis is computed in different forms depending on 
constraints of parameters.
In the system of polynomial equations given as an inverse kinematics problem,
by expressing the coordinates of the end-effector as parameters, 
then, by computing CGS from the polynomial system, 
we obtain the Gr\"obner basis where the 
coordinates of the end-effector are expressed in terms of parameters.
When moving the robot, the coordinates of the end-effector are 
substituted into the Gr\"obner basis corresponding to the segment
in which the coordinates satisfy constraints on the parameters, 
then solved the configuration of the joints.
This allows us to solve the system of polynomial equations 
immediately without computing Gr\"obner basis when 
the robot is actually in motion.

Furthermore, we have focused on quantifier elimination with CGS
(CGS-QE)~\cite{fuk-iwa-sat2015}.
CGS-QE is a QE method based on CGS, and it is said to be 
effective when the constraints have mainly equality constraints.
When we use CGS to solve inverse kinematics problems for 
the above purposes, the CGS-QE method also allows us to 
verify the existence of a solution to the inverse kinematics
problem.
Then, if the given inverse kinematic problem is determined to be 
feasible, 
it is possible to immediately obtain a solution to the 
inverse kinematic problem without Gr\"obner basis computation.

With these motivations, we have proposed an inverse 
kinematics solver that verifies the existence 
of a solution to the inverse kinematics problem by the CGS-QE method, 
and efficiently finds a feasible solution using CGS \cite{ota-ter-mik2021}.
Our solver uses ``preprocessing steps \cite[Algorithm 1]{ota-ter-mik2021}'' 
to configure the solver
before the startup of the manipulator, that is, 
we eliminate segments without real points and,
if the input system is a non-zero dimensional ideal, 
we find a trivial root that makes the input system zero-dimensional.
Then, when the manipulator is running, the solver uses
``main steps \cite[Algorithm 2]{ota-ter-mik2021}'' to determine the existence of feasible solutions 
and compute them. 
However, in the proposed algorithm, the preprocessing steps were 
performed manually.

The main contribution of this paper is the extension of 
our previous solver~\cite{ota-ter-mik2021} in two ways.
The first is that the computation of the preprocessing steps 
is completely automated. 
The procedures in the previous work were refined into an 
algorithm that can be executed automatically.
The second is the extension of the solver to 
path planning (trajectory planning) in two ways.

Trajectory planning is a computation in which the path 
along which the manipulator (the end-effector) is to be moved is given in advance, 
and the configuration of the joints is determined at each time so that 
the position of the end-effector changes as a function of time along that path.
Trajectory planning also considers the manipulator's kinematic constraints
to determine the configuration of the joints at each time.

Our extension of the solver to trajectory planning is as follows. 
The first method iteratively solves the inverse kinematics problem
along a path using the proposed method described above.
In the second method, the path is represented by a function of a parameter.
Feasibility of the inverse kinematics problem is determined using
the CGS-QE method within a given time range.
It determines whether the entire trajectory falls within the manipulator's 
feasible region before the manipulator moves. If the trajectory planning is feasible,
we solve the inverse kinematics problem sequentially along the path.

This paper is organized as follows. In \cref{sec:inverse-kinematics-problem}, 
the inverse kinematics problem for the 3-DOF manipulator is formulated for the use of 
Gr\"obner basis computation.
In \cref{sec:cgs-qe}, CGS, CGS-QE method, and a method of real root counting are reviewed. 
In \cref{sec:inverse-kinematics},  
an extension of a solver for inverse kinematics problem based on the CGS-QE method is proposed.
In \cref{sec:trajectory}, trajectory planning methods based on
the CGS-QE method are presented.
In \cref{sec:remark}, conclusions and future research topics are discussed.

\section{Inverse Kinematics of a 3-DOF Robot Manipulator}
\label{sec:inverse-kinematics-problem}

\begin{figure}[t]
    \centering
    \includegraphics[scale=0.15]{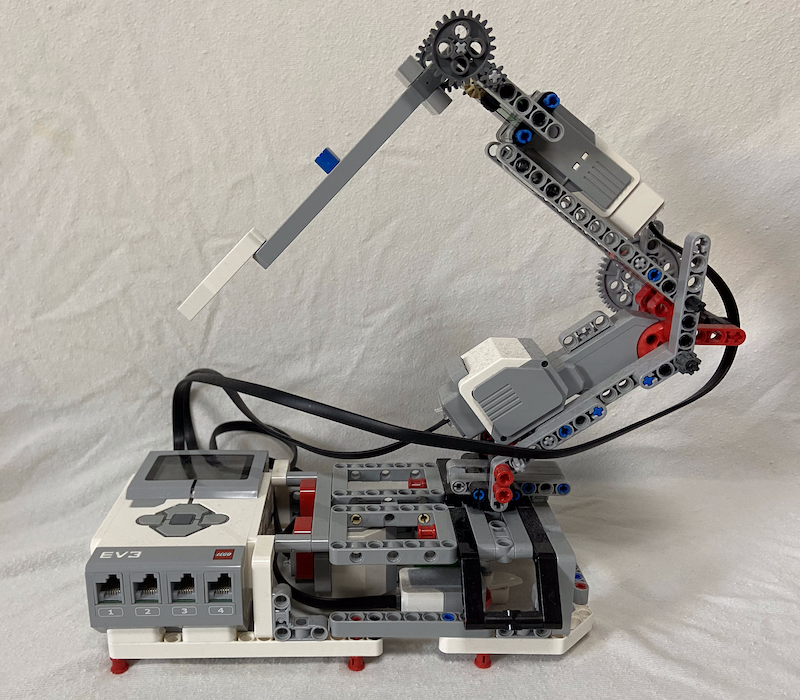}
    \caption{A 3-DOF manipulator built with EV3.}
    \label{fig:ev3-photo}
\end{figure}

In this paper, as an example of a 3-DOF manipulator, 
one built with 
LEGO\textsuperscript{\textregistered}
MINDSTORMS\textsuperscript{\textregistered} EV3 Education%
\footnote{LEGO and MINDSTORMS are trademarks of the LEGO Group.}
(henceforth abbreviated to EV3) is used (\cref{fig:ev3-photo}).
The EV3 kit is equipped with large and small motors, 
optical, touch, gyro sensors, and a computer called 
``EV3 Intelligent Brick.''
A GUI-based development environment is provided, and 
development environment with Python, Ruby, C, and Java are 
also available.

The components of the manipulator is shown in 
\cref{fig:ev3-components}. 
\begin{figure}[t]
    \centering
    \scalebox{0.6}{
    \begin{tikzpicture}[scale=0.2,
        cross/.style={
            path picture={ 
                \draw[black]
                (path picture bounding box.south east) -- 
                (path picture bounding box.north west) 
                (path picture bounding box.south west) -- 
                (path picture bounding box.north east);
            }
        },
        axis/.style={
            inner sep=0pt, 
            outer sep=0pt,
            minimum width=0.15cm,
            minimum height=0.15cm
        },
        dot/.style={
            path picture={ 
                \draw[fill, color=black]
                circle (0.02cm);
            }
        },
        ]
        \tikzset{
            ev3/.pic = {
                Joint 0
                \node [ground, rotate=90,
                label={[xshift=0.5cm, yshift=-1cm] Joint 0}](j0) at (0,1) {};

                \node [
                label={[xshift=0cm, yshift=-1.2cm] Joint 1}](j1) at (-2,1) {};
                \node [draw, diamond, aspect = 2, rotate=90,
                label={[xshift=1.1cm, yshift=-0.3cm] $\theta_1$}](j1) at (-2,1) {};
                \draw[->,>=stealth] (-1.8,0.5) to [out=45,in=315] (-1.8,1.5);

                \node [draw, circle,
                label={[xshift=0cm, yshift=-1.25cm] Joint 2}](j2) at (-4,1) {};

                \node [draw, circle, above left =2cm of j2](j3) {};
                \node [draw, circle, above left =2cm of j2,
                label={[xshift=0cm, yshift=0.4cm] Joint 3}](j3) {};

                \node [draw, circle, left = 2cm of j3,
                label={[xshift=0cm, yshift=0.4cm] Joint 4}](j4) {};
                \node [
                label={[xshift=0.8cm, yshift=-0.5cm] $\theta_4$}]() at (-8,3) {};
                \draw[<-,>=stealth] (-8.5,2.75)  to [out=45,in=135] (-7.5,2.75);

                \node [draw, circle, left = 2cm of j4,
                label={[xshift=0cm, yshift=-1.2cm] Joint 5}](j5) {};

                \node [draw, circle, above = 2cm of j5](j6) {};
                \node [draw, circle, above = 2cm of j5,
                label={[xshift=1cm, yshift=-0.4cm] Joint 6}](j6) {};

                \node [
                label={[xshift=0.8cm, yshift=-0.5cm] $\theta_7$}] at (-12.5,5.5) {};
                \draw[<-,>=stealth] (-13.1,5.1)  to [out=45,in=135] (-12.1,5.1);
                \node [draw, circle, left = 2cm of j6,
                label={[xshift=0cm, yshift=0.4cm] Joint 7}](j7) {};

                \coordinate [left = 2cm of j7] (j8) {};
                \node [draw, circle, left = 2cm of j7, label={[xshift=0cm, yshift=0.4cm] Joint 8}](j8) {};
                \coordinate [above left = 1cm of j8](j811) {};
                \coordinate [left = 1cm of j811](j812) {};
                \coordinate [below left = 1cm of j8](j821) {};
                \coordinate [left = 1cm of j821](j822) {};

                \draw (j1) -- (j0)
                node [midway, label={[xshift=0.2cm,yshift=0.2cm] Link 0}] {};;

                \draw (j1) -- (j2)
                node [midway, label={[xshift=0.1cm,yshift=0.2cm] Link 1}] {};;

                \draw (j2) -- (j3)
                node [midway, label={[xshift=-1cm,yshift=-0.75cm] Link 2}] {};;

                \draw (j3) -- (j4)
                node [midway, label={[xshift=0cm,yshift=-0.75cm] Link 3}] {};;

                \draw (j4) -- (j5)
                node [midway, label={[xshift=0cm,yshift=-0.75cm] Link 4}] {};;

                \draw (j5) -- (j6)
                node [midway, label={[xshift=1cm,yshift=-0.25cm] Link 5}] {};;

                \draw (j6) -- (j7)
                node [midway, label={[xshift=0cm,yshift=-0.75cm] Link 6}] {};;

                \draw (j7) -- (j8)
                node [midway, label={[xshift=0cm,yshift=-0.75cm] Link 7}] {};;
                \draw (j8) -- (j811);
                \draw (j811) -- (j812);
                \draw (j8) -- (j821);
                \draw (j821) -- (j822);

                \node [
                    draw, circle, axis, cross,
                    label={[xshift=0.2cm, yshift=-0.5cm] $y_0$},
                    label={[xshift=-0.5cm, yshift=0.1cm] $\rmSigma_0$}
                ](s0) at (0.5,7){};
                \draw [->] (s0) -- ++(180:0.5cm) node [left] {$z_0$};
                \draw [->] (s0) -- ++(90:0.5cm) node [above] {$x_0$};

                \node [
                    draw, circle, axis, cross,
                    label={[xshift=0.2cm, yshift=-0.5cm] $y_1$},
                    label={[xshift=-0.5cm, yshift=0.1cm] $\rmSigma_1$}
                ](s0) at (-2,7){};
                \draw [->] (s0) -- ++(180:0.5cm) node [left] {$z_1$};
                \draw [->] (s0) -- ++(90:0.5cm) node [above] {$x_1$};

                \node [
                    draw, circle, axis, dot,
                    label={[xshift=0.2cm, yshift=0.1cm] $z_2$},
                    label={[xshift=-0.5cm, yshift=-0.2cm] $\rmSigma_2$}
                ](s2) at (-4,7){};
                \draw [->] (s2) -- ++(135:0.5cm) node [above] {$x_2$};
                \draw [->] (s2) -- ++(225:0.5cm) node [below] {$y_2$};

                \node [
                    draw, circle, axis, dot,
                    label={[xshift=0.2cm, yshift=-0.1cm] $z_3$},
                    label={[xshift=-0.5cm, yshift=-0.7cm] $\rmSigma_3$}
                ](s0) at (-5.75,7){};
                \draw [->] (s0) -- ++(180:0.5cm) node [left] {$x_3$};
                \draw [->] (s0) -- ++(270:0.5cm) node [below] {$y_3$};

                \node [
                    draw, circle, axis, dot,
                    label={[xshift=0.2cm, yshift=-0.1cm] $z_4$},
                    label={[xshift=-0.5cm, yshift=-0.7cm] $\rmSigma_4$}
                ](s0) at (-7.75,7){};
                \draw [->] (s0) -- ++(180:0.5cm) node [left] {$x_4$};
                \draw [->] (s0) -- ++(270:0.5cm) node [below] {$y_4$};

                \node [
                    draw, circle, axis, dot,
                    label={[xshift=0.2cm, yshift=-0.5cm] $z_5$},
                    label={[xshift=-0.5cm, yshift=0.1cm] $\rmSigma_5$}
                ](s0) at (-10.25,0.5){};
                \draw [->] (s0) -- ++(180:0.5cm) node [left] {$y_5$};
                \draw [->] (s0) -- ++(90:0.5cm) node [above] {$x_5$};

                \node [
                    draw, circle, axis, dot,
                    label={[xshift=0.2cm, yshift=-0.1cm] $z_6$},
                    label={[xshift=-0.5cm, yshift=-0.7cm] $\rmSigma_6$}
                ](s0) at (-10.25,7){};
                \draw [->] (s0) -- ++(180:0.5cm) node [left] {$x_6$};
                \draw [->] (s0) -- ++(270:0.5cm) node [below] {$y_6$};

                \node [
                    draw, circle, axis, dot,
                    label={[xshift=0.2cm, yshift=-0.1cm] $z_7$},
                    label={[xshift=-0.5cm, yshift=-0.7cm] $\rmSigma_7$}
                ](s0) at (-12.625,7){};
                \draw [->] (s0) -- ++(180:0.5cm) node [left] {$x_7$};
                \draw [->] (s0) -- ++(270:0.5cm) node [below] {$y_7$};

                \node [
                    draw, circle, axis, dot,
                    label={[xshift=0.2cm, yshift=-0.1cm] $z_8$},
                    label={[xshift=-0.5cm, yshift=-0.7cm] $\rmSigma_8$}
                ](s0) at (-14.875,7){};
                \draw [->] (s0) -- ++(180:0.5cm) node [left] {$x_8$};
                \draw [->] (s0) -- ++(270:0.5cm) node [below] {$y_8$};
            }         
        }

        \draw (0,0) pic {ev3};
    \end{tikzpicture}
    }
    \caption{Components and the coordinate systems of the manipulator.}
    \label{fig:ev3-components}
\end{figure}
The manipulator has eight links (segments) and eight joints connected alternatively.
A link fixed to the bottom is called Link $0$, and the other links are 
numbered as Link $1,\dots,7$ towards the end-effector. 
For $j=1,\dots,7$, the joint connecting Links $j-1$ and $j$ is called Joint $j$.
The foot of Link $0$ on the ground is called Joint $0$, and the end-effector 
is called Joint $8$.
Due to the circumstances of the appropriate coordinate transformation described below, 
Joints $1$ and $2$ overlap, and Link $1$ does not exist either.
(Note that by setting joint parameters appropriately, 
the consistency of coordinate transformation is maintained even for such a combination of links and segments.)
Joints $1(=2),4,7$ are revolute joints, while the other joints are fixed. 

At Joint $i$, according to a modified Denavit-Hartenberg convention \cite{sic-sci-vil-ori2008}, 
the coordinate system $\rmSigma_i$ is defined as follows (\cref{fig:ev3-components}). 
The origin is located at Joint $i$, and the $x_i$, $y_i$ and $z_i$ axes are 
defined as follows (in \cref{fig:ev3-components}, the positive axis pointing upwards and downwards is denoted by ``$\odot$'' and ``$\otimes$'', respectively):
\begin{itemize}
    \item The $z_j$ axis is chosen along with the axis of rotation of Joint $j$.
    \item The $x_{j-1}$ axis is selected along with the common normal to axes $z_{j-1}$ to
    $z_j$.
    \item The $y_j$ axis is chosen so that the present coordinate system is right-handed.
\end{itemize}

Note that the above definition of axes may have ambiguity. 
For the current manipulator, if the axes $z_i$ and $z_{i-1}$ are parallel, 
there are infinite ways to take the $x_i$ axis. Thus, in this case, the $x_i$ axis 
is defined as follows.
\begin{itemize}
    \item In the coordinate system $\rmSigma_0$, define the axes $x_0,y_0,z_0$ 
    like those in $\rmSigma_1$ as depicted in \cref{fig:ev3-components}.
    Also, in the coordinate system $\rmSigma_8$, define the axes $x_8,y_8,z_8$ 
    like those in $\rmSigma_7$, respectively.
    \item In the coordinate system $\rmSigma_i$ ($i=2,\dots,7$), since the origin is located 
    on Joint $i$, define the $x_i$ axis to overlap Link $i$.
\end{itemize}

For analyzing the motion of the manipulator, we define a map between the \emph{joint space}
and the \emph{configuration space} or \emph{operational space}. For a joint space, since we have revolute joints 
$1,4,7$, their angles $\theta_1,\theta_4,\theta_7$, respectively, are located 
in a circle $S^1$, we define the joint space as $\mathcal{J}=S^1\times S^1\times S^1$.
For a configuration space, let $(x,y,z)$ be the end-effector position located in 
$\R^3$ and then define the configuration space as $\mathcal{C}=\R^3$. Thus, we consider a map 
$f:\mathcal{J}\longrightarrow\mathcal{C}$. The forward kinematic problem is to find 
the position of the end-effector in $\mathcal{C}$ for the given configuration of 
the joints in $\mathcal{J}$, while the inverse kinematic problem is 
to find the configuration of the joints in $\mathcal{J}$ which enables the 
given position of the end-effector in $\mathcal{C}$. 
We first solve the forward kinematic problem for formulating the inverse kinematic problem. 

Let $a_i$ be the distance between axes $z_{i-1}$ and $z_i$,
$\alpha_i$ the angle between axes $z_{i-1}$ and $z_i$ with respect to 
the $x_i$ axis, 
$d_i$ the distance between the axes $x_{i-1}$ and $x_i$, 
and $\theta_i$ be the angle between the axes $x_{i-1}$ and $x_i$
with respect to the $z_i$ axis. Then, the coordinate transformation matrix
${}^{i-1}T_i$ from the coordinate system $\rmSigma_{i}$ to $\rmSigma_{i-1}$
is expressed as in \cref{fig:i-1Ti}.

\begin{figure}[t]
    \centering
    \begin{align*}
        {}^{i-1} T_i &=
        \begin{pmatrix}
            1 & 0 & 0 & a_i \\
            0 & 1 & 0 & 0 \\
            0 & 0 & 1 & 0 \\
            0 & 0 & 0 & 1
        \end{pmatrix}
        \begin{pmatrix}
            1 & 0 & 0 & 0 \\
            0 & \cos \alpha_i & - \sin \alpha_i & 0  \\
            0 & \sin \alpha_i & \cos \alpha_i & 0 \\
            0 & 0 & 0 & 1
        \end{pmatrix}
        \begin{pmatrix}
            1 & 0 & 0 & 0 \\
            0 & 1 & 0 & 0 \\
            0 & 0 & 1 & d_i \\
            0 & 0 & 0 & 1
        \end{pmatrix}
        \begin{pmatrix}
            \cos \theta_i & -\sin \theta_i & 0 & 0 \\
            \sin \theta_i & \cos \theta_i & 0 & 0 \\
            0 & 0 & 1 & 0 \\
            0 & 0 & 0 & 1 
        \end{pmatrix}
        \\
        &= 
        \begin{pmatrix}
            \cos \theta_i & -\sin \theta_i & 0 & a_i \\
            \cos \alpha_i \sin \theta_i & \cos \alpha_i \cos \theta_i & -\sin \alpha_i & -d_i \sin \alpha_i   \\
            \sin \alpha_i \sin \theta_i & \sin \alpha_i \cos \theta_i & \cos \alpha_i & d_i \cos \alpha_i \\
            0 & 0 & 0 & 1
        \end{pmatrix}
        ,
    \end{align*}
    \caption{The transformation matrix ${}^{i-1}T_i$.}
    \label{fig:i-1Ti}
\end{figure}
where the joint parameters $a_i$, $\alpha_i$, $d_i$ and $\theta_i$ 
are shown in \Cref{tab:ev3-parameters} (note that the unit
of $a_i$ and $d_i$ is [mm]).
\begin{table}[t]
    \caption{Joint parameters for EV3.}
    \label{tab:ev3-parameters}
    \centering
    \begin{tabular}{c|cccc}
            $i$ & $a_i$ (mm) & $\alpha_i$ & $d_i$ (mm) & $\theta_i$  \\ 
            \hline
            1 & 0 & 0 & 80 & $\theta_1 $ \\
            2 & 0 & $\pi/2$ & 0 & $\pi/4$ \\
            3 & ${88}$ & 0 & 0 &$\pi/4 $ \\
            4 & ${24} $ & 0 & 0 & $\theta_4$\\
            5 & ${96} $ & 0 & 0 & $-\pi/2$ \\
            6 & ${16}$ & 0 & 0 & $\pi/2$\\
            7 & ${40} $ & 0 & 0 &  $\theta_7$\\  
            8 & ${120} $ & 0 & 0 &0  \\
            \hline
    \end{tabular}
\end{table}
The transformation matrix $T$ 
from the coordinate system $\rmSigma_8$ to $\rmSigma_0$
is calculated as 
$T={}^{0}T_1{}^{1}T_2{}^{2}T_3{}^{3}T_4{}^{4}T_5{}^{5}T_6{}^{6}T_7{}^{7}T_8$,
where ${}^{i-1}T_i$ is expressed as in \cref{fig:i-1Tiseries}.

\begin{figure}
    \begin{align*}
        \label{eq:i-1Tiseries}
        {}^{0}T_1  
        &= 
        \left(
        \begin{array}{cccc}
        \cos \theta_1 & -\sin \theta_1 & 0 & 0 \\
        \sin \theta_1 & \cos \theta_1 & 0 & 0  \\
        0& 0 & 1 & 80 \\
        0 & 0 & 0 & 1
        \end{array}
        \right),&
        {}^{1}T_2
        &=
        \left(
        \begin{array}{cccc}
        \frac{\sqrt{2}}{2}  & -\frac{\sqrt{2}}{2}  & 0 & 0 \\
        0 & 0 & -1 & 0 \\
        \frac{\sqrt{2}}{2} & \frac{\sqrt{2}}{2} & 0 & 0  \\
        0 & 0 & 0 & 1
        \end{array}
        \right),&
        {}^{2}T_3 
        &= 
        \left(
        \begin{array}{cccc}
        \frac{\sqrt{2}}{2} & -\frac{\sqrt{2}}{2} & 0 & {88} \\
        \frac{\sqrt{2}}{2} & \frac{\sqrt{2}}{2} & 0 & 0  \\
        0& 0 & 1 & 0 \\
        0 & 0 & 0 & 1
        \end{array}
        \right),
        \\
        {}^{3}T_4
        &= 
        \left(
        \begin{array}{cccc}
        \cos \theta_4 & -\sin \theta_4 & 0 & {24} \\
        \sin \theta_4 & \cos \theta_4 & 0 & 0  \\
        0& 0 & 1 & 0 \\
        0 & 0 & 0 & 1
        \end{array}
        \right),&
        {}^{4}T_5
        &= 
        \left(
        \begin{array}{cccc}
        0 & 1 & 0 & {96} \\
        -1 & 0 & 0 & 0 \\
        0& 0 & 1 & 0 \\
        0 & 0 & 0 & 1
        \end{array}
        \right),&
        {}^{5}T_6  
        &= 
        \left(
        \begin{array}{cccc}
        0 & -1 & 0 & {16} \\
        1 & 0 & 0 & 0  \\
        0& 0 & 1 & 0 \\
        0 & 0 & 0 & 1
        \end{array}
        \right),\\
        {}^{6}T_7
        &= 
        \left(
        \begin{array}{cccc}
            \cos \theta_7 & -\sin \theta_7 & 0 & {40}  \\
            \sin \theta_7 & \cos \theta_7 & 0 & 0 \\
        0& 0 & 1 & 0 \\
        0 & 0 & 0 & 1
        \end{array}
        \right),&
        {}^{7}T_8  
        &= 
        \left(
        \begin{array}{cccc}
        1 & 0 & 0 & {120}  \\
        0 & 1 & 0 & 0  \\
        0& 0 & 1 & 0 \\
        0 & 0 & 0 & 1
        \end{array}
        \right).
    \end{align*}
    \caption{The transformation matrix ${}^{i-1}T_i$ ($i=1,\dots,8$).}
    \label{fig:i-1Tiseries}
\end{figure}

Then, the position $(x,y,z)$ 
of the end-effector with respect to the coordinate system $\rmSigma_0$ is
expressed as 
\begin{equation}
    \label{eq:ev3-forward-kinematics}
    \begin{split}
        x &= -120\cos{\theta_1}\cos{\theta_4}\sin{\theta_7}+16\cos{\theta_1}\cos{\theta_4}-120\cos{\theta_1}\sin{\theta_4}\cos{\theta_7}\\
        &\qquad -136{\cos}\theta_1{\sin}\theta_4+44{\sqrt{2}}\cos{\theta_1},  \\
        y &= -120\sin{\theta_1}\cos{\theta_4}\sin{\theta_7}+16\sin{\theta_1}\cos{\theta_4}-120\sin{\theta_1}\sin{\theta_4}\cos{\theta_7}\\
        &\qquad -136\sin{\theta_1}\sin{\theta_4}+44{\sqrt{2}}\sin{\theta_1},  \\
        z &= 120\cos{\theta_4}\cos{\theta_7}+136{\cos}{\theta_4}-120\sin{\theta_4}\sin{\theta_7}+16{\sin}{\theta_4}
         +104+44{\sqrt{2}}.
    \end{split}
\end{equation}

The inverse kinematics problem comes down to solving \cref{eq:ev3-forward-kinematics}
for $\theta_1$, $\theta_4$, $\theta_7$.
By substituting trigonometric functions $\cos\theta_i$ and $\sin\theta_i$ with variables as
$c_i=\cos\theta_i,s_i=\sin\theta_i,$
subject to $c_i^2+s_i^2=1$, \cref{eq:ev3-forward-kinematics} is transferred to a system of 
polynomial equations:
\begin{equation}
    \label{eq:inverse-kinematic-equations}
    \begin{split}
        f_1 &= 120c_1c_4s_7-16c_1c_4+120c_1s_4c_7+136c_1s_4-44{\sqrt{2}}c_1 + x = 0, \\
        f_2 &= 120s_1c_4s_7-16s_1c_4+120s_1s_4c_7+136s_1s_4-44{\sqrt{2}}s_1 + y = 0, \\
        f_3 &= -120c_4c_7-136c_4+120s_4s_7-16s_4-104-44{\sqrt{2}} + z = 0, \\
        f_4 &= s_1^2+c_1^2-1=0, \quad
        f_5 = s_4^2+c_4^2-1=0, \quad
        f_6 = s_7^2+c_7^2-1=0.
    \end{split}
\end{equation}

\section{Real Quantifier Elimination Based on CGS}
\label{sec:cgs-qe}

Equations~\eqref{eq:ev3-forward-kinematics} and \eqref{eq:inverse-kinematic-equations} show that solving the inverse kinematic problem 
for the given system can be regarded as a real quantifier elimination of 
a quantified formula
\begin{multline}
    \label{eq:inverse-kinematic-qe-formula}
    \exists{c_1}\exists{s_1}\exists{c_4}\exists{s_4}\exists{c_7}
    \exists{s_7} \\
    (f_1=0 \land f_2=0 \land f_3=0 \land f_4=0 \land f_5=0 \land f_6=0),
\end{multline}
with $x,y,z$ as parameters.

In this section, we briefly review an algorithm of real quantifier elimination 
based on CGS, the CGS-QE algorithm,
by Fukasaku et al. \cite{fuk-iwa-sat2015}.
Two main tools play a crucial role in the algorithm: one is CGS, 
and another is real root counting, or counting the number of real roots of a system of polynomial equations.
Note that, in this paper, we only consider equations in the quantified formula. 

Hereafter, let $R$ be a real closed field, $C$ be the algebraic closure of $R$, and
$K$ be a computable subfield of $R$. 
This paper considers $R$ as the field of 
real numbers $\R$, $C$ as the field of complex numbers $\C$, and $K$ as 
the field of rational numbers $\Q$.
Let $\Xbar$ and $\Abar$ denote variables $X_1,\dots,X_n$ and $A_1,\dots,A_m$,
respectively, and $T(\Xbar)$ be the set of the monomials
which consist of variables in $\Xbar$. For an ideal $I\subset K[\Xbar]$, let
$V_R(I)$ and $V_C(I)$ be the affine varieties of $I$ in $R$ or $C$, respectively, 
satisfying that 
$V_R(I)=\{\cbar\in R^n\mid \mbox{$\forall f(\Xbar)\in I$: $f(\cbar)=0$}\}$
and
$V_C(I)=\{\cbar\in C^n\mid \mbox{$\forall f(\Xbar)\in I$: $f(\cbar)=0$}\}$.

\subsection{CGS}

For the detail and algorithms on CGS, see Fukasaku et al. \cite{fuk-iwa-sat2015} or references therein. 
In this paper, the following notation is used. 
Let $\succ$ be an admissible term order. For a polynomial $f\in K[\Abar,\Xbar]$ with
a term order $\succ$ on $T(\Xbar)$, we regard $f$ as a polynomial in 
$(K[\Abar])[\Xbar]$, which is the ring of polynomials with $\Xbar$ as variables and
coefficients in $(K[\Abar])$ such that $\Abar$ is regarded as parameters.
Given a term order $\succ$ on $T(\Xbar)$,
$\lt(f)$, $\lc(f)$ and $\lm(f)$ denotes the leading term,
the leading coefficient, and the leading monomial, respectively, satisfying that
$\lt(f)=\lc(f)\lm(f)$ with $\lc(f)\in K[\Abar]$ and $\lm\in T(\Xbar)$
(we follow the notation by Cox et al. \cite{cox-lit-osh2015}). 

\begin{definition}[Algebraic Partition and Segment]
    Let $S\subset C^m$ for $m\in\N$. A finite set 
    $\{\mathcal{S}_1,\dots,\mathcal{S}_t\}$ of nonempty subsets of $S$ 
    is called an algebraic partition of $S$ if it satisfies the following 
    properties:
    \begin{enumerate}
        \item $S=\bigcup_{k=1}^t\mathcal{S}_k$.
        \item For $k\ne j\in\{1,\dots,t\}$, $\mathcal{S}_k\cap\mathcal{S}_j=\emptyset$.
        \item For $k\in\{1,\dots,t\}$, $\mathcal{S}_k$ is expressed as
        $\mathcal{S}_k=V_C(I_1)\setminus V_C(I_2)$ for some ideals $I_1,I_2\subset K[\Abar]$.
    \end{enumerate}
    Furthermore, each $\mathcal{S}_k$ is called a segment.
\end{definition}

\begin{definition}[Comprehensive Gr\"obner System (CGS)]
    \label{def:CGS}
    Let $S\subset C^m$ and $\succ$ be a term order on $T(\Xbar)$. 
    For a finite subset $F\subset K[\Abar,\Xbar]$, a finite set
    $\mathcal{G}=\{(\mathcal{S}_1,G_1),\dots,(\mathcal{S}_t,G_t)\}$ is 
    called a Comprehensive Gr\"obner System (CGS) of $F$ over $\mathcal{S}$
    with parameters $\Abar$ with respect to $\succ$ if it satisfies
    the following:
    \begin{enumerate}
        \item For $k\in\{1,\dots,t\}$, $G_k$ is a finite subset of $K[\Abar,\Xbar]$.
        \item The set $\{\mathcal{S}_1,\dots,\mathcal{S}_t\}$ is an algebraic partition of 
        $\mathcal{S}$.
        \item For each $\cbar\in\mathcal{S}_k$, 
        $G_k(\bar{c},\bar{X})=\{g(\bar{c},\bar{X})\mid g(\Abar,\bar{X})\in G_k\}$
        is a Gr\"obner basis of the ideal 
        $\ideal{F(\bar{c},\bar{X})}\subset C[\bar{X}]$ with respect to $\succ$, where
        $F(\bar{c},\bar{X})=\{f(\bar{c},\bar{X})\mid f(\Abar,\bar{X}) \in F\}$.
        \item For each $\bar{c}\in S_k$, any $g\in G_k$ satisfies that 
        $\big(\lc(g)\big)(\bar{c})\neq 0$.
    \end{enumerate}
    Furthermore, if each $G_k(\bar{c},\bar{X})$ is a minimal or the reduced Gr\"obner basis,
    $\mathcal{G}$ is called a minimal or the reduced CGS, respectively. 
    In the case $\mathcal{S}=C^m$, the words ``over $\mathcal{S}$'' may be omitted.
\end{definition}

\subsection{Real Root Counting}
\label{sec:real-root-counting}

Let $I \subset {K}[\Xbar]$ be a zero-dimensional ideal. Then,
the quotient ring $K[\Xbar]/I$ is regarded as a finite-dimensional vector
space over $K$ \cite{cox-lit-osh2005}; let $\{v_1,\dots,v_d\}$ be its basis.
For $h\in K[\Xbar]/I$ and $i,j$ satisfying $1\le i,j\le d$,
let $\theta_{h,i,j}$ be a linear transformation defined as
\[
    \begin{array}{cccc}
        {\theta_{h,i,j}} : & {K}[\bar{X}]/I & {\longrightarrow} & {K}[\bar{X}]/I \\
        & \rotatebox{90}{$\in$} & & \rotatebox{90}{$\in$} \\
        & f & \mapsto & hv_iv_jf
    \end{array}
    .
\]
Let $q_{h,i,j}$ be the trace of $\theta_{h,i,j}$ and $M_h^I$ be a symmetric matrix
such that its $(i,j)$-th element is given by $q_{h,i,j}$.
Let $\chi_h^I(X)$ be the characteristic polynomial of $M_h^I$, and $\sigma(M_h^I)$,
called the signature of $M_h^I$, 
be the number of positive eigenvalues of $M_h^I$ minus the number of negative 
eigenvalues of $M_h^I$. Then, we have the following theorem on the real root counting
\cite{bec-woe1994,ped-roy-szp1993}.

\begin{theorem}[The Real Root Counting Theorem]
    We have
    \[
        \sigma(M_h^I)=\#(\{\cbar\in V_R(I)\mid h(\cbar)>0\})
        -\#(\{\cbar\in V_R(I)\mid h(\cbar)<0\}).
    \]
\end{theorem}

\begin{corollary}
    \label{cor:signature-real-roots}
    $\sigma(M_1^I)=\#(V_R(I))$.
\end{corollary}

Since we only consider a quantified formula with equations, as in 
\Cref{eq:inverse-kinematic-qe-formula}, we omit properties of the real root 
counting related to quantifier elimination of quantified formula with inequalities or inequations (for detail, see Fukasaku et al. \cite{fuk-iwa-sat2015}).

\subsection{CGS-QE Algorithm}
\label{sec:cgs-qe-alg}


The CGS-QE algorithm accepts the following quantified formula given as 
\begin{gather*}
    \exists \Xbar (f_1(\Abar,\Xbar)=0  \land  \cdots  \land  f_\mu(\Abar,\Xbar)=0  \land
    p_1(\Abar,\Xbar)>0  \land  \cdots  \land  p_\nu(\Abar,\Xbar)>0  \land \\
    q_1(\Abar,\Xbar)\ne 0  \land  \cdots  \land  q_\xi(\Abar,\Xbar)\ne 0),
    \\
    f_1,\dots,f_\mu,p_1,\dots,p_\nu,q_1,\dots,q_\xi \in\Q[\Abar,\Xbar]\setminus\Q[\Abar],
\end{gather*}
then outputs an equivalent quantifier-free formula.
Note that, in this paper, we give a quantified formula only with equations as shown 
in \Cref{eq:inverse-kinematic-qe-formula}.
The algorithm is divided into several algorithms. 
The main algorithm is called \textbf{MainQE}, and sub-algorithms are called
\textbf{ZeroDimQE} and \textbf{NonZeroDimQE} for the case that
the ideal generated by the component of the CGS is zero-dimensional or 
positive dimensional, respectively.
(For a complete algorithm description, see Fukasaku et al. \cite{fuk-iwa-sat2015}.)

In the real root counting, we need to calculate $\sigma(M_h^I)$ as in 
\Cref{sec:real-root-counting}. This calculation is executed using 
the following property \cite{wei1998} derived from Descartes' rule of signs.
Let $M$ be a real symmetric matrix of dimension $d$ and 
$\chi(X)$ be the characteristic polynomial of $M$ of degree $d$,
expressed as
\begin{equation}
    \label{eq:chix}
    \small
    \begin{split}    
        \chi(\lambda) = \lambda^d+a_{d-1}\lambda^{d-1}+{\ldots}+a_0,\
        \chi(-\lambda) = (-1)^d\lambda^d+b_{d-1}\lambda^{d-1}+{\ldots}+b_0.
    \end{split}
\end{equation}
Note that $b_\ell=a_\ell$ if $\ell$ is even, and $b_\ell=-a_\ell$ if $\ell$
is odd. 
Let $\Lplus{\chi}$ and $\Lminus{\chi}$ be the sequence of the coefficients
in $\chi(\lambda)$ and $\chi(-\lambda)$, defined as
\begin{equation}
    \label{eq:L+-}
    \Lplus{\chi}=(1,a_{d-1},\dots,a_0),\quad
    \Lminus{\chi}=((-1)^d,b_{d-1},\dots,b_0),
\end{equation}
respectively.
Furthermore, let $\Lbarplus{\chi}$ and $\Lbarminus{\chi}$ be the sequences
defined by removing zero coefficients in $\Lplus{\chi}$ and
$\Lminus{\chi}$, respectively, and let
\begin{equation}
    \label{eq:S+-}
    \begin{split}
        \Splus{\chi} &= (\text{the number of sign changes in $\Lbarplus{\chi}$}), \\
        \Sminus{\chi} &= (\text{the number of sign changes in $\Lbarminus{\chi}$}).
    \end{split}
\end{equation}
Then, we have the following.

\begin{lemma}
    Let $\Splus{\chi}$ and $\Sminus{\chi}$ be defined as in \cref{eq:S+-}.
    Then, we have
    \[
            \Splus{\chi} = \#(\{c\in R\mid c>0\land \chi(c)=0\}),\,
            \Sminus{\chi} = \#(\{c\in R\mid c<0\land \chi(c)=0\}).
    \]
\end{lemma}

\begin{corollary}
    \label{cor:signature-cgs-qe}
    Let $\Splus{\chi}$ and $\Sminus{\chi}$ be defined as in \cref{eq:S+-},
    and
    $I$ be a zero-dimensional ideal and
    $M_1^I$ be a matrix defined as in \Cref{sec:real-root-counting}.
    Then, we have 
    \begin{equation}
        \label{eq:signature}
        {\#}(V_{R}(I))=\sigma(M^I_1)\Leftrightarrow\Splus{\chi}\ne\Sminus{\chi}.
    \end{equation}
\end{corollary}

\begin{remark}
    As shown below, most of our inverse kinematic computation uses up to the real root counting part of the CGS-QE algorithm. The part of the algorithm that eliminates quantified variables and obtains conditions on the parameters is used only to verify the feasibility of the inverse kinematic solution for the given path (see \cref{sec:path-planning-with-cgs-qe}).
\end{remark}

\section{Solving the Inverse Kinematic Problem}
\label{sec:inverse-kinematics}

This section shows a method for solving the inverse kinematic problem 
in \cref{eq:inverse-kinematic-equations}.
Specifically, for the coordinates of the end-effector that are given as 
$(x,y,z)=(x_0,y_0,z_0)\in\R^3$, 
determine the feasibility of the configuration of the end-effector 
with the CGS-QE method.
If the configuration of the end-effector is feasible, then 
compute $c_1,s_1,c_4,s_4,c_7,s_7$ by solving \cref{eq:inverse-kinematic-equations}, 
and compute the angle $\theta_1,\theta_4,\theta_7$ of Joint $1,4,7$, respectively, as 
\begin{equation}
    \label{eq:calculate-joint-angles}
    \theta_1=\arctan(s_1/c_1),\,\theta_4=\arctan(s_4/c_4),\,\theta_7=\arctan(s_7/c_7).    
\end{equation}

The computation is executed as follows, summarized as \Cref{alg:solve-ikp-point}.
For \Cref{alg:solve-ikp-point}, $f_1,\dots,f_6$ in \cref{eq:inverse-kinematic-equations},
variables $\Xbar=(c_1,s_1,c_4,s_4,c_7,s_7)$ parameters $\Abar=(x,y,z)$, and a position 
of the end-effector $\p=(x_0,y_0,z_0)$ are given.
(For optional arguments, see \cref{rem:solve-ikp-point}.)
\begin{enumerate}
    \item 
    Compute CGS of $\ideal{f_1,\dots,f_6}\subset\R[\Abar,\Xbar]$ with an appropriate monomial order.
    Let 
    \begin{equation}
        \label{eq:cF}
        \cF=\{(\cS_1,G_1),\dots,(\cS_t,G_t)\}     
    \end{equation}
    be the computed CGS.
    Assume that the segment $\cS_k$ is represented as 
    \begin{equation}
        \label{eq:cS}
        \cS_k=V_{\C}(I_{k,1})\setminus V_{\C}(I_{k,2}),\quad
        I_{k,1}=\ideal{F_{k,1}},\quad
        I_{k,2}=\ideal{F_{k,2}},
    \end{equation}
    where $F_{k,1},F_{k,2}\subset\R[\Abar]$.
    \item From $\cF$, eliminate 
    $(\cS,G)\in\cF$ satisfying that
    $\cS\cap\R^3=\emptyset$ and that are easily detected.
    Re-arrange indices as 
    $\cF'=\{(\cS_1,G_1),\dots,(\cS_t,G_\tau)\}$.
    See \Cref{sec:Generate-Real-CGS} for detail.    
    \item For $(x_0,y_0,z_0)$,  
    choose $(\cS_k,G_k)\in\cF'$ satisfying that $(x_0,y_0,z_0)\in \cS_k$.
    Let 
    \begin{equation}
            \label{eq:substitutedGB}
            G=\{g_1,\dots,g_\rho\},
    \end{equation}
    be $G_k$ with substituting $(x_0,y_0,z_0)$ for $(x,y,z)$.
    \item For $G$ in \cref{eq:substitutedGB}, determine if $\ideal{G}$ is zero-dimensional.
    For the case $\ideal{G}$ is not zero-dimensional, see \Cref{sec:Solve-IKP-NonZeroDim}.
    \item If $\ideal{G}$ is zero-dimensional, calculate the number of real roots of 
    \begin{equation}
        \label{eq:ik-equation-substituted}
        g_1=\cdots=g_\rho=0.    
    \end{equation}
    See \Cref{sec:Count-Real-Roots} for detail.
    \item If the system of polynomial equations \eqref{eq:ik-equation-substituted}
    has real roots, calculate approximate roots with a numerical method.
    If the system has more than one set of real roots, we accept the first set 
    of roots that the solver returns.
    \item By \cref{eq:calculate-joint-angles}, calculate joint angles 
    $\theta_1,\theta_4,\theta_7$.
\end{enumerate}


\begin{algorithm}[t]
    \caption{Solving the inverse kinematic problem}
    \label{alg:solve-ikp-point}
  \begin{algorithmic}[1]
    \Require $F=\{f_1,\dots,f_6\}$: \cref{eq:inverse-kinematic-equations} for the inverse kinematic problem,
    $\cV=\{c_1,s_1,c_4,s_4,c_7,s_7\}$: variables,
    $\cP=\{x,y,z\}$: parameters,
    $\p=(x_0,y_0,z_0)$: a position of the end-effector to be placed,
    $\cF$ (optional): a CGS of $\ideal{F}$ or the output of
    \textproc{Generate-Real-CGS($\cF$,$\cP$)}
    (\Cref{alg:Generate-Real-CGS}) where $\cF$ is a CGS of $\ideal{F}$,
    $\CallGenerateRealCGS = \{\TRUE\mid\FALSE\}$ (optional): whether one wish to call 
    \textproc{Generate-Real-CGS} (\Cref{alg:Generate-Real-CGS}) or not;
    \Ensure $\Theta=\{\theta_1,\theta_4,\theta_7\}$: joint angles of a solution of the inverse kinematic problem, or $\Theta=\emptyset$
    if there are no solution or an infinite number of solutions;
    \Function{Solve-IKP-Point}{$F$, $\cV$, $\cP$, $\p$, $\cF$, \CallGenerateRealCGS}
        \If {$\cF=\emptyset$}
            \State Compute a CGS of $\ideal{F}$ as $\cF=\{(\cS_1,G_1),\dots,(\cS_t,G_t)\}$;
            \label{line:alg:solve-ikp-point:cgs}
        \EndIf
        \If {$\CallGenerateRealCGS=\TRUE$}
            \State $\cF'\gets\text{\textproc{Generate-Real-CGS}($\cF,\cV$)}$;
            \label{line:alg:solve-ikp-point:generate-real-cgs}
            \Comment{See \Cref{sec:Generate-Real-CGS} (\Cref{alg:Generate-Real-CGS})}
        \Else\ {$\cF'\gets\cF$;}
        \EndIf
        \State Choose $(\cS_k,G_k)$ from the CGS $\cF'$ satisfying  
        $\p\in\cS_k$; 
        \State $G'\gets\{g\in G\mid \text{$x\gets x_0,y\gets y_0,z\gets z_0$ in $g$}\}$
        \State $\sigma\gets\textproc{Count-Real-Roots}(G')$;
        \Comment{See \Cref{sec:Count-Real-Roots} (\Cref{alg:Count-Real-Roots})}
        \If {$\sigma=\textrm{``\FAIL''}}$
            \State $\Theta\gets\textproc{Solve-IKP-NonZeroDim($G'$)}$;
            \Comment{See \Cref{sec:Solve-IKP-NonZeroDim}
            (\Cref{alg:Solve-IKP-NonZeroDim})}
        \ElsIf{$\sigma=0$}\ $\Theta\gets\emptyset$;
        \Else 
            \label{line:alg:solve-ikp-point:modified-mainqe}
            \State $S\gets 
            (\text{real solutions of $g_1=\cdots=g_\rho=0$ in \cref{eq:ik-equation-substituted}})$; 
            \State $\Theta\gets(\text{joint angles obtained by \cref{eq:calculate-joint-angles}})$;
        \EndIf
        \State \textbf{return} $\Theta$;
    \EndFunction
  \end{algorithmic}
\end{algorithm}

\begin{remark}
    \label{alg:solve-ikp-point}
    We see that \cref{alg:solve-ikp-point} outputs $\Theta=\{\theta_1,\theta_4,\theta_7\}$ or 
    $\Theta=\emptyset$ correctly, as follows.
    After computing the CGS $\cF$, some segments without real points are eliminated optionally, 
    resulting in $\cF'$. Then, a pair of a segment and the accompanying Gr\"obner basis 
    $(\cS_k,G_k)$ is chosen, satisfying that $\p\in\cS_k$. After defining $G'$ by substituting 
    parameters $(x,y,z)$ in $g\in G$ with $\p$, The number of real roots of polynomial 
    equations $\{g'=0\mid g'\in G'\}$ is counted by \Cref{alg:Count-Real-Roots}, and 
    it returns $\sigma$.
    In the case $\sigma=0$, this means that there are 
    no real roots in $\{g'=0\mid g'\in G'\}$, thus $\emptyset$ is output.
    In the case $\sigma=\textrm{``FAIL''}$, $G'$ is investigated by 
    \cref{alg:Solve-IKP-NonZeroDim} and a value of $\emptyset$ or $\Theta$ is returned,
    which becomes the output of this algorithm.
    Finally, in the case $\sigma>0$, real solutions of $\{g'=0\mid g'\in G'\}$ are calculated as 
    $\Theta$, which becomes the output of this algorithm. This finishes the computation.
\end{remark}

\begin{remark}
    \label{rem:solve-ikp-point}
    In \Cref{alg:solve-ikp-point}, 
    it is also possible to calculate the GCS $\cF$ or $\cF'$ 
    (in which some segments without real points are eliminated)
    first and then given to the algorithm.
    The arguments $\cF$ and \CallGenerateRealCGS\ in the function \textproc{Solve-IKP-Point} 
    are optional.
    Furthermore, if $\cF'$ is given to \textproc{Solve-IKP-Point}, the variable
    \CallGenerateRealCGS\ is set \TRUE.
    Pre-computing the CGS before executing \Cref{alg:solve-ikp-point} would make the algorithm more efficient, especially when repeatedly solving the same problem
    (see \Cref{ex:time-based-path-trajectory}).
\end{remark}

\subsection{Removing a Segment not Existing in $\R^3$}
\label{sec:Generate-Real-CGS}

In the inverse kinematic problem, since the 
parameters consist of $x,y,z$ in \cref{eq:inverse-kinematic-equations},
the segments in the algebraic partition corresponding to the CGS 
$\cF$ in \cref{eq:cF} exist in $\C^3$.
However, since only real values of $x,y,z$ are used in 
solving the inverse kinematic problem,
if a segment $\cS_k$ in \cref{eq:cS} do not exist in $\R^3$, then it
can be ignored.
Thus, by investigating generators in 
$F_{k,1}$ and $F_{k,2}$ in \cref{eq:cS}, 
we remove some $\cS_k$ that satisfies 
$\cS_k\cap\R^3=\emptyset$ and that is 
easily detected, as follows, 
summarized as \Cref{alg:Generate-Real-CGS}.

\begin{enumerate}
    \item Let $f\in F_{k,1}$. If $f$ is a univariate polynomial and $\deg f=2$, calculate the 
    discriminant $\disc{f}$ of $f$. If $\disc{f}<0$, then remove $(\cS_k,G_k)$.
    \item If $f$ is a univariate polynomial and $\deg f\ge 3$, calculate the number of real roots 
    of $f$ by the Sturm's method. If the number of real roots of $f$ is equal to 
    $0$, then remove $(\cS_k,G_k)$.
    \item \label{step:find-trivial-roots} 
    Let $(x_0,y_0,z_0)$ be a root of $f\in F_{k,1}$ as many coordinates as possible are $0$.
    Assume that there exists $f_0\in F_{k,1}$ with only the real root $(x_0,y_0,z_0)$ 
    (for detecting $f_0$ satisfying this property, see below).
    \item If there exists $g\in F_{k,1}$ satisfying that $g(x_0,y_0,z_0)$ is
    a nonzero constant,
    then we see that 
    $(x_0,y_0,z_0)\not\in\cS_k\cap\R^3$, thus remove $(\cS_k,G_k)$.    
    \item If all $h\in F_{k,2}$ satisfies $h(x_0,y_0,z_0)=0$,
    then we see that $(x_0,y_0,z_0)\not\in\cS_k\cap\R^3$, thus remove $(\cS_k,G_k)$.
\end{enumerate}

\begin{algorithm}
    \caption{Removing a segment which does not exist in $\R^3$}
    \label{alg:Generate-Real-CGS}
  \begin{algorithmic}[1]
    \Require $\cF=\{(\cS_1,G_1),\dots,(\cS_t,G_t)\}$: a CGS,
    $\cP$: parameters
    \Ensure $\cF'=\{(\cS_1,G_1),\dots,(\cS_\tau,G_\tau)\}$: a CGS with organized numbering 
    in which segments those do not exist in $\R^3$ are removed;
    \Function{Generate-Real-CGS}{$\cF$, $\cP$}
        \State{$\text{Undecided}\gets\text{True}$};
        \ForEach{$(\cS,G)\in\cF$}
            \State $(x_0,y_0,z_0)\gets(x,y,z)$;
            \ForEach{$f\in F_1$ where $\cS=V_C(I_1)\setminus V_C(I_2)$, $I_1=\ideal{F_1}$ and $I_2=\ideal{F_2}$} 
                \If {$f$ is a univariate polynomial} 
                    \If {$\deg f\ge 3$}
                        \State $\#\text{RealRoots}\gets$ (the number of reall roots of $f$ computed with the Sturm's method);
                        \If {$\#\text{RealRoots}=0$} {$\text{Undecided}\gets\text{False}$; break;}
                        \EndIf
                    \ElsIf {$\deg f = 2$}
                        \If {$\disc{f}=0$} {$\text{Undecided}\gets\text{False}$; break;}
                        \EndIf
                    \EndIf
                \Else \
                $(x_0,y_0,z_0)\gets\textproc{Find-Trivial-Roots}(f,(x_0,y_0,z_0))$;
                \Comment{\Cref{alg:find-trivial-roots}}
                \If {$(x_0,y_0,z_0)=\emptyset$} 
                    $\text{Undecided}\gets\text{False}$; break;
                \EndIf
                \EndIf
            \EndFor
            \If {$(x_0,y_0,z_0)=\emptyset$} 
            $\text{Undecided}\gets\text{False}$;
            \Else
                \ForEach {$g\in F_1$}
                    \If {$g(x_0,y_0,z_0)$ is a nonzero constant}
                        $\text{Undecided}\gets\text{False}$; break;
                    \EndIf
                \EndFor
                    \If {for all $g\in F_2$ $g(x_0,y_0,z_0)=0$}
                        $\text{Undecided}\gets\text{False}$;
                    \EndIf
            \EndIf
            \If {$\text{Undecided}=\text{True}$}
                {$\cF'\gets\cF'\cup\{(\cS,G)\}$;}
            \EndIf
        \EndFor       
        \State {Renumber indices of $(\cS,G)$ in $\cF'$ as 
        $\cF'=\{(\cS_1,G_1),\dots,(\cS_\tau,G_\tau)\}$;}
        \State \textbf{return} $\cF'$
    \EndFunction
  \end{algorithmic}
\end{algorithm}

In Step~\ref{step:find-trivial-roots} above, 
we find $(x_0,y_0,z_0)$, a root of $f\in F_{k,1}$ as many coordinates as possible are $0$,
along with $f_0$ which has $(x_0,y_0,z_0)$ only the real root, as follows.
For the purpose, we find $f$ 
with the terms of the degree with respect to each parameter $x,y,z$ is even, 
expressed as 
\begin{equation}
    \label{eq:f-even-degrees}
    f=a+\sum_{(p,q,r)\in\Z_{\ge0}^3 \setminus\{(0,0,0)\}} a_{p,q,r} x^{2p}y^{2q}z^{2r},\quad a\in\R,\quad a_{p,q,r}\ne 0.
\end{equation}
We see that $f$ of the form as in \cref{eq:f-even-degrees} may have the following property.
\begin{enumerate}
    \item If $a\ne 0$ and the signs of $a$ and $a_{p,q,r}$ ($a_{p,q,r}\ne 0$) are the same, 
    then $f$ does not have a real root. 
    \item If $a=0$ and the signs of $a$ and $a_{p,q,r}$ ($a_{p,q,r}\ne 0$) are the same, 
    then $f$ has a root that the parameters appearing in $f$ equals $0$.
    Let $(x_0,y_0,z_0)$ be $(x,y,z)$ with the variable appearing in $f$ set to $0$.
\end{enumerate}

\begin{example}
    Examples of polynomials of the form as in \cref{eq:f-even-degrees} satisfying properties in above.
    \begin{enumerate}
        \item A polynomial with $a\ne 0$ and the signs of $a$ and $a_{p,q,r}$ ($a_{p,q,r}\ne 0$) are the same: $f_1(x,y,z)=2x^2y^4+z^2+3=0$ does not have a real root.
        \item A polynomial with $a=0$ and the signs of $a$ and $a_{p,q,r}$ ($a_{p,q,r}\ne 0$) are the same: $f_2(x,y,z)=-2x^2y^4-z^2=0$ has a trivial real root $x=y=z=0$.
    \end{enumerate}
\end{example}
By \Cref{alg:find-trivial-roots}, we find a polynomial that has no real roots
or $f_0$ that has only the real root $(x_0,y_0,z_0)$ with as many coordinates as possible are $0$.

\begin{algorithm}[t]
    \caption{Find a roots as many coordinates as possible are $0$}
    \label{alg:find-trivial-roots}
  \begin{algorithmic}[1]
    \Require $f\in\R[\Abar]$, $(x_0,y_0,z_0)$: $x_0\in\{x,0\},y_0\in\{y,0\},z_0\in\{z,0\}$;
    \Ensure $(x_0,y_0,z_0)$: $x_0\in\{x,0\},y_0\in\{y,0\},z_0\in\{z,0\}$ or $\emptyset$;
    \Function{Find-Trivial-Roots}{$f$, $(x_0,y_0,z_0)$}
        \If {$f$ is expressed as in \cref{eq:f-even-degrees}}
            \If {$a\ne 0$}
                \If {the signs of $a$ and $a_{p,q,r}$ are the same}
                    $(x_0,y_0,z_0)\gets\emptyset$;
                \EndIf
            \ElsIf {the signs of $a$ and $a_{p,q,r}$ are the same}
                \If {$x$ appears in $f$}
                    $x_0\gets 0$
                \ElsIf {$y$ appears in $f$}
                    $y_0\gets 0$
                \ElsIf {$z$ appears in $f$}
                    $z_0\gets 0$
                \EndIf
           \EndIf
        \EndIf
        \State \textbf{return} $(x_0,y_0,z_0)$;
    \EndFunction
  \end{algorithmic}
\end{algorithm}

\begin{remark}
    We see that \cref{alg:find-trivial-roots} finds a polynomial of 
    the form of \eqref{eq:f-even-degrees}
    that has no real roots or $f_0$ that has only the real root 
    $(x_0,y_0,z_0)$ with as many coordinates as possible are 0,
    as follows. 
    If $f$ is the form of \eqref{eq:f-even-degrees} with $a\ne 0$, 
    investigate if signs of $a$ and the other non-zero coefficients are 
    the same. If the signs are the same, $f$ does not have a real root, 
    and the algorithm returns $\emptyset$.
    On the other hand, if $f$ is the form of \eqref{eq:f-even-degrees} with $a=0$
    and signs of the other non-zero coefficients are the same, 
    $f$ has a unique root with $x=0$, $y=0$ or $z=0$. 
    Then, $x_0$, $y_0$ or $z_0$ are replaced with $0$ if corresponding variables
    appears in $f$.
\end{remark}

\begin{remark}
    We see that \cref{alg:Generate-Real-CGS} outputs a CGS $\cF$ with some 
    segments without real points eliminated, as follows. 
    Let $\cS_k$, $I_{k,1}$, $I_{k,2}$, $F_{k,1}$ and $F_{k,2}$ be as in 
    \cref{eq:cS}.
    If $f\in F_{k,1}$ is a univariate polynomial, 
    real roots are counted using the discriminant (if $\deg f=2$) or 
    Sturm's method (if $\deg f\ge 3$).
    Thus, if $f$ is a univariate polynomial with no real toot, then $\cS_k$ has no real point.
    Next, for $f\in F_{k,1}$ expressed as in \eqref{eq:f-even-degrees},
    \Cref{alg:find-trivial-roots} reports that there exists $f\in F_{k,1}$ that does not have a real root or finds a root $(x_0,y_0,z_0)$ with as many coordinates as possible are $0$.
    \begin{enumerate}
        \item If $f\in F_{k,1}$ has no real root, then $\cS_k$ has no real point.
        \item If there exists a root $(x_0,y_0,z_0)$ with as many coordinates as possible are $0$, since the form of the input polynomial in \Cref{alg:find-trivial-roots} is as in \Cref{eq:f-even-degrees}, we see that $(x_0,y_0,z_0)$ is a root of $f_0\in F_{k,1}$ that has no other real roots. 
        We examine if $(x_0,y_0,z_0)\in \cS_k=V_{\C}(I_{k,1})\setminus V_{\C}(I_{k,2})$.
        If there exists $g\in F_{k,1}$ satisfying that $g(x_0,y_0,z_0)$ is
        a nonzero constant, then $(x_0,y_0,z_0)\not\in V_{\C}(I_{k,1})$, thus 
        $(x_0,y_0,z_0)\not\in\cS_k$.
        Futhermore, if all $h\in F_{k,2}$ satisfies $h(x_0,y_0,z_0)=0$,
        $(x_0,y_0,z_0)\in V_{\C}(I_{k,2})$, thus $(x_0,y_0,z_0)\not\in\cS_k$.
    \end{enumerate}
\end{remark}

\begin{remark}
    Even without \Cref{alg:Generate-Real-CGS}, it is possible to eventually remove segments 
    that do not have a real point in \Cref{alg:solve-ikp-point}.
    However, it may be possible to improve the efficiency of solving the inverse kinematic problem 
    while iterating \Cref{alg:solve-ikp-point} by providing a CGS that has previously removed segments that do not have real number points using \Cref{alg:Generate-Real-CGS}
    (see \Cref{ex:time-based-path-trajectory}).
\end{remark}

\subsection{Calculating the Number of Real Roots}
\label{sec:Count-Real-Roots}

Calculating the number of real roots in 
\cref{eq:inverse-kinematic-equations} is based on 
Algorithm MainQE in the CGS-QE method \cite{fuk-iwa-sat2015}.
While the original algorithm computes constraints on parameters 
such that the equations have a real root, 
the parameters are substituted with the coordinates of the end-effector, 
thus the number of real roots is calculated as follows,
summarized as \Cref{alg:Count-Real-Roots}.
\begin{enumerate}
    \item Let $G$ be the Gr\"obner basis $G$ in \cref{eq:substitutedGB}.
    Determine if $\ideal{G}$ is zero-dimensional. If 
    $\ideal{G}$ is not zero-dimensional, apply 
    computation in \Cref{sec:Solve-IKP-NonZeroDim}.
    \item Calculate a real symmetric matrix $M_1^{\ideal{G}}$
    (for its definition, see \Cref{sec:real-root-counting}).
    \item By \Cref{cor:signature-real-roots}, calculate the number of real roots of 
    $\{g=0\mid g\in G\}$ by calculating $\sigma(M_1^{\ideal{G}})$.
\end{enumerate}

\begin{algorithm}[t]
    \caption{Calculating the number of real roots \cite{fuk-iwa-sat2015}}
    \label{alg:Count-Real-Roots}
  \begin{algorithmic}[1]
    \Require $G$: a Gr\"obner basis as in \cref{eq:substitutedGB}
    \Ensure $\sigma$: the number of real roots of $\{g=0\mid g\in G\}$; 
    In the case $\ideal{G}$ is not zero-dimensional, return $\sigma=\textrm{``\FAIL''}$;
    \Function{Count-Real-Roots}{$G$}
        \If {$\ideal{G}$ is zero-dimensional} 
            \State $\sigma\gets\sigma(M_{1}^{\ideal{G}})$;
            \Comment{Calculated by \Cref{cor:signature-real-roots}}
        \Else\
            $\sigma\gets\textrm{``\FAIL''}$;
            \Comment{
            See \Cref{sec:Solve-IKP-NonZeroDim}
            }
        \EndIf
        \State \textbf{return} $\sigma$;
    \EndFunction
  \end{algorithmic}
\end{algorithm}

\begin{remark}
    For a Gr\"obner basis $G$, we see that \cref{alg:Count-Real-Roots} counts the number of real roots of $\{g=0\mid g\in G\}$ if $\ideal{G}$ is zero-dimensional. If $\ideal{G}$ is zero-dimensional, then the number of real roots is calculated by \cref{cor:signature-real-roots}. On the other hand, $\ideal{G}$ is not zero-dimensional, it returns ``\FAIL.''
\end{remark}

\subsection{Calculation for Non-Zero Dimensional Ideals}
\label{sec:Solve-IKP-NonZeroDim}

Our previous studies \cite{ota-ter-mik2021} have shown that, for $G$ in \cref{eq:substitutedGB},
there exists a case that $\ideal{G}$ is not zero-dimensional. 
In the case $x_0=y_0=0$, $c_1^2+s_1^2-1\in G$ and the corresponding segment
$\cS$ satisfies $\cS=V_{\C}(I_1)\setminus V_{\C}(I_2)$, $I_1=\ideal{x,y}$.
(Note that, in this case, the segment $\cS$ is different from the one in which the most 
feasible end-effector positions exist.)
This means that the points in $V_{\R}(I_1)$ satisfy $x=y=0$, and 
the end-effector is located on the $z$-axis in the coordinate system $\rmSigma_0$. 
In this case, $\theta_1$, the angle of Joint 1 is not uniquely determined. 
Then, by putting $\theta_1=0$ (i.e., $c_1=1,s_1=0$) in $g\in G$,
we obtain a new system of polynomial equations $G'$ which satisfies that $\ideal{G'}$ is
zero-dimensional, and, by solving a new system of polynomial equations
$\{g'=0\mid g'\in G'\}$,
 a solution to the inverse kinematic problem is obtained.

Based on the above observations, for $G$ in \cref{eq:substitutedGB}, in the case, $\ideal{G}$ is not zero-dimensional, we perform the following calculation, summarized as \Cref{alg:Solve-IKP-NonZeroDim}.

\begin{enumerate}
    \item It is possible that $G$ has a polynomial $g_0=s_1^2+c_1^2-1$.
    If such $g_0$ exists, 
    define 
    \[
        G'=\{g\in G\setminus\{g_0\}\mid \textrm{substitute $s_1\gets 1$ and $c_1\gets 0$ in $g$}\}.  
    \]
    \item For newly defined $G'$, apply \Cref{alg:Count-Real-Roots} for testing if 
    $G'$ is zero-dimensional. 
    If $G'$ is zero-dimensional, calculate the number of real
    roots of the system of equations
    \begin{equation}
        \label{eq:nonzerodim-G}
        g'_1=\cdots=g'_{\rho}=0,
    \end{equation} 
    where $g'_1,\dots,g'_{\rho}\in G$.
    \item If the number of real roots of \cref{eq:nonzerodim-G} is positive, then compute approximate real roots and put then into $\Theta$.
\end{enumerate}

\begin{algorithm}[t]
    \caption{Computing real roots for non-zero dimensional ideal}
    \label{alg:Solve-IKP-NonZeroDim}
  \begin{algorithmic}[1]
    \Require $G$: a Gr\"obner basis of non-zero dimensional ideal
    \Ensure $\Theta=\{\theta_1,\theta_4,\theta_7\}$: joint angles of a solution of the inverse kinematic problem, or $\Theta=\emptyset$
    if there are no solution or an infinite number of solutions;
    \Function{Solve-IKP-NonZeroDim}{$G$}
        \State $G'\gets\emptyset$;
        \ForEach{$g\in G$}
            \If {$g\ne s_1^2+c_1^2-1$}
                $G'\gets G'\cup \{g\}$;
            \EndIf
        \EndFor
        \ForEach {$g'\in G'$}
        $s_1\gets 1$; $c_1\gets 0$;
        \EndFor
        \State $\sigma\gets\textproc{Count-Real-Roots($G'$)}$;
        \Comment{See \Cref{sec:Count-Real-Roots} (\Cref{alg:Count-Real-Roots})}
        \If {$\sigma=0$ or ``\FAIL''}
            $\Theta\gets\emptyset$;
        \Else
            \State $S\gets(\text{real solutions of $\{g'=0\mid g\in G'\}$})$;
            \State $\Theta\gets(\text{joint angles obtained by \cref{eq:calculate-joint-angles}})$;
        \EndIf
        \State \textbf{return} $\Theta$;
    \EndFunction
  \end{algorithmic}
\end{algorithm}

\begin{remark}
    For a Gr\"obner basis $G$ of non-zero dimensional ideal, we see that 
    \cref{alg:Solve-IKP-NonZeroDim} outputs 
    $\Theta=\{\theta_1,\theta_4,\theta_7\}$ or 
    $\Theta=\emptyset$ correctly, as follows.
    $G'$ is calculated as $G'=\{g\in G\mid g \ne s_1^2+c_1^2-1\}$.
    Then, for $g'\in G'$, $s_1\leftarrow 0$ and $c_1\leftarrow 1$.
    The number of real roots of polynomial 
    equations $\{g'=0\mid g'\in G'\}$ is counted by \Cref{alg:Count-Real-Roots}, and 
    it returns $\sigma$.
    In the case $\sigma=0$, this means that there are 
    no real roots in $\{g'=0\mid g'\in G'\}$, thus $\emptyset$ is output.
    In the case $\sigma=\textrm{``FAIL''}$, 
    further computation is cancelled and $\emptyset$ is output.
    Finally, in the case $\sigma>0$, real solutions of $\{g'=0\mid g'\in G'\}$ are calculated as 
    $\Theta$, which becomes the output of this algorithm. 
\end{remark}

\begin{remark}
    Note that \Cref{alg:Generate-Real-CGS,alg:find-trivial-roots,alg:Solve-IKP-NonZeroDim} 
    correspond to ``preprocessing steps (Algorithm 1)'' in our previous solver 
    \cite{ota-ter-mik2021}.
    In our previous solver, except for the computation of the CGS, 
    ``the rest of computation was executed by hand'' \cite[Sect.\ 4]{ota-ter-mik2021}.
\end{remark}

\subsection{Experiments}
\label{sec:inverse-kinematics-experiments}

We have implemented and tested the above inverse kinematics solver
\cite{ev3-cgs-qe-ik-2}.
An implementation was made on the computer algebra system Risa/Asir \cite{nor2003}.
Computation of CGS was executed with the implementation by Nabeshima \cite{nab2018}.
The computing environment is as follows:
Intel Xeon Silver 4210 3.2 GHz, 
RAM 256 GB,
Linux Kernel 5.4.0,
Risa/Asir Version 20230315.

Test sets for the end-effector's position were the same as those used 
in the tests of our previous research \cite{hor-ter-mik2020,ota-ter-mik2021}.
The test sets consist of 10 sets of 100 random end-effector positions  
within the feasible region; thus, 1000 random points were given.
The coordinates of the position were given as rational numbers with the magnitude of the denominator less than 100.
For solving a system of polynomial equations numerically, 
computer algebra system PARI-GP 2.3.11 \cite{pari2.13.1} was used
in the form of a call from Risa/Asir.
In the test, we have used pre-calculated CGS of \cref{eq:inverse-kinematic-equations}
(originally, to be calculated in Line~\ref{line:alg:solve-ikp-point:cgs} of 
\Cref{alg:solve-ikp-point}). The computing time of CGS was approximately 62.3 sec.

\Cref{tab:inverse-kinematics-point} shows the result of experiments.
In each test, `Time' is the average computing time (CPU time), rounded at the 5th decimal place. 
`Error' is the average of the absolute error, or the 2-norm distance of the end-effector from the randomly given position to the calculated position with the configuration of the computed joint angles $\theta_1,\theta_4,\theta_7$. 
The bottom row, `Average' shows the average values in each column of the 10 test sets.

The average error of the solution was approximately $1.63\times 10^{-12}$ [mm].
Since the actual size of the manipulator is approximately 100 [mm], computed solutions with the present method seem sufficiently accurate.
Comparison with data in our previous research shows that the current result 
is more accurate than our previous result 
($1.982\times 10^{-9}$ [mm] \cite{ota-ter-mik2021} and $4.826\times 10^{-11}$ [mm] \cite{hor-ter-mik2020}). Note that the software used for solving equations in the current experiment  
differs from the one used in our previous experiments; this could have affected the results.

The average computing time for solving the inverse kinematic problem was approximately
100 [ms]. 
Comparison with data in our previous research shows that the current result 
is more efficient than our previous result
($540$ [ms] \cite{ota-ter-mik2021} and $697$ [ms] \cite{hor-ter-mik2020}, 
measured in the environment of Otaki et al. \cite{ota-ter-mik2021}).
However, systems designed for real-time control using Gr\"obner basis computation have achieved computation times of 10 [ms] order \cite{uch-mcp2011,uch-mcp2012}.
Therefore, our method may have room for improvement (see \cref{sec:remark}).

\begin{table}[t]
    \centering
    \caption{A result of inverse kinematics computation.}
    \label{tab:inverse-kinematics-point}
    \begin{tabular}{rcl}
        \hline
        Test & Time (sec.) & Error (mm) \\
        \hline
        1 & 0.1386 & $1.2428\times 10^{-12}$ \\
        2 & 0.1331 & $2.3786\times 10^{-12}$ \\
        3 & 0.1278 & $1.0845\times 10^{-12}$ \\
        4 & 0.1214 & $1.6150\times 10^{-12}$ \\
        5 & 0.1147 & $1.5721\times 10^{-12}$ \\
        6 & 0.1004 & $1.6229\times 10^{-12}$ \\
        7 & 0.0873 & $2.2518\times 10^{-12}$ \\
        8 & 0.0792 & $1.3923\times 10^{-12}$ \\
        9 & 0.0854 & $1.2919\times 10^{-12}$ \\
        10 & 0.0797 & $1.8674\times 10^{-12}$ \\
        \hline
        Average & 0.1068 & $1.6319\times 10^{-12}$ \\
        \hline
    \end{tabular}
\end{table}

\section{Path and Trajectory Planning}
\label{sec:trajectory}

In this section, we propose methods for path and trajectory planning 
of the manipulator based on the CGS-QE method.

In path planning, we calculate the configuration of the joints for moving the position of the
end-effector along with the given path. 
In trajectory planning, we calculate the position (and possibly its velocity and acceleration)
of the end-effector as a function of time series depending on constraints on the velocity 
and acceleration of the end-effector and other constraints.

In \Cref{sec:time-based-path-trajectory},
we make a trajectory of the end-effector to move it along 
a line segment connecting two different points in $\R^3$
with considering constraints on the velocity and acceleration of 
the end-effector. 
Then, by the repeated use of inverse kinematics solver proposed 
in \Cref{sec:inverse-kinematics}, we calculate a series of 
configuration of the joints. 
In \Cref{sec:path-planning-with-cgs-qe}, for the path
of a line segment expressed with a parameter, 
we verify that by using the CGS-QE method, moving the end-effector along the path
is feasible for a given range of the parameter, then perform trajectory planning
as explained in the previous subsection.

\subsection{Path and Trajectory Planning for a Path Expressed as a Function of Time}
\label{sec:time-based-path-trajectory}

Assume that the end-effector of the manipulator moves along a line segment 
from the given initial to the final position as follows.
\begin{itemize}
    \item $\p_d=\T(x,y,z)$: current position of the end-effector,
    \item $\p_0=\T(x_0,y_0,z_0)$: the initial position of the end-effector,
    \item $\p_f=\T(x_f,y_f,z_f)$: the final position of the end-effector,
\end{itemize}
where $\p_d,\p_0,\p_f\in\R^3$ and $x_0,y_0,z_0,x_f,y_f,z_f$ are constants satisfying 
$x_0\ne x_f$, $y_0\ne y_f$, $z_0\ne z_f$.
Then, with a parameter $s\in[0,1]$, $\p_d$ is expressed as
\begin{equation}
    \label{eq:p_d}
    \p_d=\p_0(1-s)+\p_f s.    
\end{equation}
Note that the initial position $\p_0$ and the final position $\p_f$ 
corresponds to the case of $s=0$ and $1$ in \cref{eq:p_d}, respectively.

Then, we change the value of $s$ with a series of time $t$.
Let $T$ be a positive integer.
For $t\in[0,T]$,
set $s$ as a function of $t$ as $s=s(t)$ 
satisfying that $s\in [0,1]$.
Let $\sdot$ and $\sddot$ be the first and the second derivatives of $s$, respectively. 
(Note that $\sdot$ and $\sddot$ corresponds to the speed and the acceleration of 
the end-effector, respectively.)

Let us express $s(t)$ as a polynomial in $t$.
At $t=0$, the end-effector is stopped at $\p_0$.
Then, accelerate and move the end-effector along with a line segment for a short while.
After that, slow down the end-effector and, at $t=T$, stop it at $\p_f$.
We require the acceleration at $t=0$ and $T$ equals $0$ for smooth starting and stopping.
Then, $s(t)$ becomes a polynomial of degree 5 in $t$ \cite{lyn-par2017}, as follows.
Let 
\begin{equation}
    \label{eq:s(t)}
    s(t) = \frac{a_4T}{5}\left(\frac{t}{T}\right)^5 + \frac{a_3T}{4}\left(\frac{t}{T}\right)^4 + \frac{a_2T}{3}\left(\frac{t}{T}\right)^3 + \frac{a_1T}{2}\left(\frac{t}{T}\right)^2 + a_0 t,
\end{equation}
where $a_4,a_3,a_2,a_1,a_0\in\R$. (Note that, for $s(0)=0$, $s(t)$ does not 
have a constant term.)
Then, we have
\begin{equation}
    \label{eq:s-derivatives}
    \begin{split}
        \sdot(t) &= a_4\left(\frac{t}{T}\right)^4 + a_3\left(\frac{t}{T}\right)^3 +  a_2\left(\frac{t}{T}\right)^2 +  a_1\left(\frac{t}{T}\right) + a_0, \\
        \sddot(t) &= \frac{4a_4}{T}\left(\frac{t}{T}\right)^3 + \frac{3a_3}{T}\left(\frac{t}{T}\right)^2 + \frac{2a_2}{T}\left(\frac{t}{T}\right) + \frac{a_1}{T}.    
    \end{split}    
\end{equation}
By the constraints $s(0)=\sdot(0)=\sddot(0)=0$, $s(T)=1$, $\sdot(T)=\sddot(T)=0$, 
we see that $a_0=a_1=0$ and $a_3,a_4,a_5$ satisfy the following system of linear equations.
\begin{equation}
    \label{eq:s-linear-system}
        20 a_2 + 15 a_3 + 12 a_4 - \frac{60}{T} = 0, \quad
        a_2 + a_3 + a_4  = 0, \quad
        2 a_2 + 3 a_3 + 4 a_4  = 0.
\end{equation}
By solving \cref{eq:s-linear-system}, we obtain 
$a_2=\frac{30}{T},a_3=-\frac{60}{T},a_4=\frac{30}{T}$. 
Thus, $s(t),\sdot(t),\sddot(t)$ become as
\begin{equation}
    \label{eq:s-solutions}
    \begin{split}
        s(t) &=  \frac{6}{T^5} t^5 - \frac{15}{T^4} t^4 + \frac{10}{T^3} t^3, \quad
        \sdot(t) = \frac{30}{T^5} t^4 - \frac{60}{T^4} t^3 + \frac{30}{T^3} t^2, \\
        \sddot(t) &= \frac{120}{T^5} t^3 - \frac{180}{T^4} t^2 + \frac{60}{T^3} t,
    \end{split}
\end{equation}
respectively.


We perform trajectory planning as follows.
For given $\p_0=\T(x_0,y_0,z_0)$, $\p_f=\T(x_f,y_f,z_f)$, $t\in[0,T]$,
calculate $s(t)$ by \cref{eq:s-solutions}.
For each value of $t$ changing as $t=0,1,\dots,T$, 
calculate $\p_d=\T(x_d,y_d,z_d)$ by \cref{eq:p_d},
then apply \Cref{alg:solve-ikp-point} with $x_d,y_d,z_d$ and calculate 
the configuration of joints $\theta_1,\theta_4,\theta_7$.

This procedure is summarized as \Cref{alg:solve-ikp-trajectory}.

\begin{algorithm}[t]
    \caption{A path and trajectory planning of the manipulator}
    \label{alg:solve-ikp-trajectory}
  \begin{algorithmic}[1]
    \Require $F=\{f_1,\dots,f_6\}$: a system of equations for the inverse kinematic problem 
    \cref{eq:inverse-kinematic-equations},
    $\cV=\{c_1,s_1,c_4,s_4,c_7,s_7\}$: variables,
    $\cP=\{x,y,z\}$: parameters,
    $\p_0=\T(x_0,y_0,z_0)$: the initial position of the end-effector in the path,
    $\p_f=\T(x_f,y_f,z_f)$: the final position of the end-effector in the path,
    $T$: a step length of the time series;
    $\cF$ (optional): a CGS of $\ideal{F}$ or the output of
    \textproc{Generate-Real-CGS($\cF$,$\cP$)}
    (\Cref{alg:Generate-Real-CGS}) where $\cF$ is a CGS of $\ideal{F}$,
    $\CallGenerateRealCGS = \{\TRUE\mid\FALSE\}$ (optional): whether one wish to call 
    \textproc{Generate-Real-CGS} (\Cref{alg:Generate-Real-CGS}) or not;
    \Ensure $L=\{\Theta_t=(\theta_{1,t},\theta_{4,t},\theta_{7,t})\mid t=1,\dots,T\}$: a series of  
    solution of the inverse kinematic problem \cref{eq:inverse-kinematic-equations};
    \Function{Compute-IKP-Trajectory}{$F$, $\cV$, $\cP$, $\p_0$, $\p_f$, $T$, $\cF$, \CallGenerateRealCGS}
        \If {$\cF=\emptyset$}
            \State Compute a CGS of $\ideal{F}$ as $\cF=\{(\cS_1,G_1),\dots,(\cS_t,G_t)\}$;
        \EndIf
        \If {$\CallGenerateRealCGS=\TRUE$}
            \State $\cF'\gets\text{\textproc{Generate-Real-CGS}($\cF,\cV$)}$;
            \label{line:alg:solve-ikp-point:generate-real-cgs}
            \Comment{See \Cref{sec:Generate-Real-CGS} (\Cref{alg:Generate-Real-CGS})}
        \Else\ {$\cF'\gets\cF$;}
        \EndIf
        \State $L\gets\emptyset$;
        \For {$t=1,\dots,T$}
            \State $s\gets \frac{6}{T^5} t^5 - \frac{15}{T^4} t^4 + \frac{10}{T^3} t^3$;
            $\p_d\gets \p_0(1-s)+\p_f$ \Comment{from \cref{eq:s-solutions,eq:p_d}, respectively};
            \State $\Theta\gets\textproc{Solve-IKP-Point($F,\cV,\cP,\p_d,\cF',\FALSE$)}$;
            \Comment{See \cref{sec:inverse-kinematics} (\Cref{alg:solve-ikp-point})}
            \If {$\Theta\ne\emptyset$}
                \State $L\gets L\cup\{\Theta\}$;
            \Else\
                \textbf{return} $L$;
            \EndIf
        \EndFor
        \State \textbf{return} $L$;
    \EndFunction
  \end{algorithmic}
\end{algorithm}

\begin{remark}
    We see that \Cref{alg:solve-ikp-trajectory} outputs a trajectory for the given path of 
    the end-effector, as follows. 
    After computing the CGS $\cF$, some segments without real points are eliminated optionally, 
    resulting in $\cF'$. Next, a trajectory of points on the given path is calculated as 
    $s(t)$ with $t=0,\dots,T$. 
    Then, for $t=0,\dots,T$, an inverse kinematic problem is solved with \Cref{alg:solve-ikp-point},
    and while the solution $\Theta$ of the inverse kinematic problem exists, a sequence of solutions $L$ is obtained.
\end{remark}

\begin{remark}
    In \Cref{alg:solve-ikp-trajectory},
    it is also possible to calculate the GCS $\cF$ or $\cF'$ 
    (in which some segments without real points are eliminated by \Cref{alg:Generate-Real-CGS})
    first and then give them to the algorithm as in the case of 
    \Cref{alg:solve-ikp-point}.
    (The specification is the same as \Cref{alg:solve-ikp-point}; see \Cref{rem:solve-ikp-point}.)
\end{remark}

\begin{example}
    \label{ex:time-based-path-trajectory}
    Let 
    $\p_0=\T(x_0,y_0,z_0)=\T(10,40,80)$,
    $\p_f=\T(x_f,y_f,z_f)=\T(40,100,20)$
    and $T=50$. 
    As the CGS corresponding to \cref{eq:inverse-kinematic-equations}, 
    $\cF$ that has already been computed in \Cref{sec:inverse-kinematics-experiments}
    is given. 
    By \Cref{alg:solve-ikp-trajectory}, a sequence $L$ of 
    the configuration of the joints $\theta_1,\theta_4,\theta_7$ 
    corresponding to each point in the trajectory of the 
    end-effector from $\p_0$ to $\p_f$
    has been obtained. The total amount of computing time (CPU time) for path and trajectory 
    planning was approximately 3.377 sec. 
    Next, we show another example by using CGS 
    $\cF'=\textproc{Generate-Real-CGS}(\cF,\cV)$, where
    $\cF$ is the same as the one used in the previous example.
    Then, the computing time (CPU time) was approximately 2.246 sec. 
    Note that computing time has been reduced by using the CGS
    with some segments not containing real points eliminated using 
    \Cref{alg:Generate-Real-CGS}.
\end{example}

\begin{remark}
    \label{rem:discontinuity}
    \Cref{alg:solve-ikp-trajectory} may cause a discontinuity in the sequence of the 
    configuration of the joints when a point on the trajectory
    gives a non-zero dimensional ideal as handled by \Cref{alg:Solve-IKP-NonZeroDim}.
    For example, assume that the trajectory has a point $\p=(0,0,z_0)$ at $t=t_0$ ($0<t_0<T$).  
    Then, at $t=t_0$, according to \Cref{alg:Solve-IKP-NonZeroDim}, $\theta_1$ is set to $0$ 
    regardless of the value of $\theta_1$ at $t=t_0-1$. This could cause $\theta_1$ to jump 
    between $t_0-1$ and $t_0$, resulting a discontinuity in the sequence of configuration of 
    Joint 1. Preventing such discontinuity in trajectory planning is one of our future challenges.
\end{remark}

\subsection{Trajectory Planning with Verification of the Feasibility of the Inverse Kinematic Solution}
\label{sec:path-planning-with-cgs-qe}

Assume that the path of the motion of the end-effector is given as 
\cref{eq:p_d} with
the initial position $\p_0$ and 
the final position $\p_f$. 
We propose a method of trajectory planning by verifying the existence 
of the solution of the inverse kinematic problem with the CGS-QE method.

In the equation of the inverse kinematic problem \eqref{eq:inverse-kinematic-equations},
by substituting parameters $x,y,z$ with the coordinates of $\p_d$ in \cref{eq:p_d},
respectively, we have the following system of polynomial equations.
\begin{equation}
    \label{eq:inverse-kinematic-equations-parameters}
    \begin{split}
      f_1 &= 120 c_1 c_4 s_7 - 16 c_1 c_4 + 120 c_1 s_4 c_7 + 136 c_1 s_4 -44 \sqrt{2} c_1 \\
      &\quad + x_0(1-s) + x_fs = 0, \\
      f_2 &= 120 s_1 c_4 s_7 - 16 s_1 c_4 + 120 s_1 s_4 c_7 + 136 s_1 s_4 -44 \sqrt{2} s_1 \\
      &\quad  + y_0(1-s) + y_fs = 0, \\
      f_3 &= -120 c_4 s_7 - 136 c_4 + 120 s_4 s_7 -16 s_4 -104 - 44 \sqrt{2} \\
      &\quad + z_0(1-s) + z_fs = 0, \\
      f_4 &= s_1^2 + c_1^2 - 1 = 0, \quad
      f_5 = s_4^2 + c_4^2 - 1 = 0, \quad
      f_6 = s_7^2 + c_7^2 - 1 = 0.
    \end{split}
\end{equation} 
Note that $x_0,y_0,z_0,x_f,y_f,z_f$ are the constants.

Equation~\eqref{eq:inverse-kinematic-equations-parameters} has a parameter $s$. 
Using the CGS-QE method, we verify \cref{eq:inverse-kinematic-equations-parameters} has real roots 
for $s\in[0,1]$. 
The whole procedure for trajectory planning is shown in \Cref{alg:solve-ikp-cgsqe}.

\begin{remark}
    In \Cref{alg:solve-ikp-cgsqe},
    it is also possible to calculate the GCS $\cF$ or $\cF'$ 
    (in which some segments without real points are eliminated by \Cref{alg:Generate-Real-CGS})
    first and then give them to the algorithm as in the case of 
    \Cref{alg:solve-ikp-point,alg:solve-ikp-trajectory}.
    (The specification is the same is \Cref{alg:solve-ikp-point,alg:solve-ikp-trajectory}; 
    see \Cref{rem:solve-ikp-point}.)
\end{remark}

\begin{algorithm}[t]
    \caption{Trajectory planning with CGS-QE method}
    \label{alg:solve-ikp-cgsqe}
  \begin{algorithmic}[1]
    \Require $F=\{f_1,\dots,f_6\}$: a system of equations for the inverse kinematic problem 
    \cref{eq:inverse-kinematic-equations},
    $\cV=\{c_1,s_1,c_4,s_4,c_7,s_7\}$: variables,
    $\cP=\{x,y,z\}$: parameters,
    $\p_0=\T(x_0,y_0,z_0)$: the initial position of the end-effector in the path,
    $\p_f=\T(x_f,y_f,z_f)$: the final position of the end-effector in the path,
    $T$: the step length of a time series;
    $\cF$ (optional): a CGS of $\ideal{F}$ or the output of
    \textproc{Generate-Real-CGS($\cF$,$\cP$)}
    (\Cref{alg:Generate-Real-CGS}) where $\cF$ is a CGS of $\ideal{F}$,
    $\CallGenerateRealCGS = \{\TRUE\mid\FALSE\}$ (optional): whether one wish to call 
    \textproc{Generate-Real-CGS} (\Cref{alg:Generate-Real-CGS}) or not;
    \Ensure $L=\{\Theta_t=(\theta_{1,t},\theta_{4,t},\theta_{7,t})\mid t=1,\dots,T\}$: a series of  
    solution of the inverse kinematic problem \cref{eq:inverse-kinematic-equations};
    \Function{Solve-IKP-Trajectory-CGS-QE}{$F$, $\cV$, $\cP$, $\p_0$, $\p_f$, $T$, $\cF$,
    \CallGenerateRealCGS}
        \If {$\cF=\emptyset$}
            \State Compute a CGS of $\ideal{F}$ as $\cF=\{(\cS_1,G_1),\dots,(\cS_t,G_t)\}$;
            \label{line:solve-ikp-cgsqe:cgs}
        \EndIf
        \If {$\CallGenerateRealCGS=\TRUE$}
            \State $\cF'\gets\text{\textproc{Generate-Real-CGS}($\cF$, $\cV$)}$;
            \label{line:solve-ikp-cgsqe:generate-real-cgs}
            \Comment{See \Cref{sec:Generate-Real-CGS} (\Cref{alg:Generate-Real-CGS})}
        \Else {}
            $\cF'\gets\cF$;
        \EndIf
        \State $M\gets\textproc{MainQE}(\cF')$;
        \label{line:solve-ikp-cgsqe:mainqe}
        \If {$[0,1]\subset M$}
            \State {$L\gets\textproc{Compute-IKP-Trajectory}(F,\cV,\cP,\p_0,\p_f,T,\cF',\FALSE)$;}
            \Comment{See \cref{sec:time-based-path-trajectory} (\Cref{alg:solve-ikp-trajectory})}
        \Else {}
            {$L\gets\emptyset$;}
        \EndIf
        \State \textbf{return} $L$;
    \EndFunction
  \end{algorithmic}
\end{algorithm}

In \Cref{alg:solve-ikp-cgsqe}, Line \ref{line:solve-ikp-cgsqe:mainqe} corresponds to
Algorithm MainQE in the CGS-QE method.  Its detailed procedure for a zero-dimensional ideal
is as follows. Let $\cF'$ be the input CGS, $(\cS,G)$ a segment, 
$M\subset\R^3$ the output. Assume that the ideal $\ideal{G}$ is 
zero-dimensional.
\begin{enumerate}
    \item In the case $G\ne\{1\}$, calculate the matrix $M_1^{\ideal{G}}$.
    \item Calculate the characteristic polynomial $\chi_{1}^{\ideal{G}}(X)$.
    \label{list:solve-ikp-cgsqe:chi}
    \item Calculate the range $M$ of parameter $s$ that makes 
    \cref{eq:inverse-kinematic-equations-parameters} has a real root as follows.
    \label{list:solve-ikp-cgsqe:simplify}
    \begin{enumerate}
        \item By \cref{eq:chix,eq:L+-}, generate the sequences of 
        coefficients $\Lplus{\chi_{1}^{\ideal{G}}}$ and $\Lminus{\chi_{1}^{\ideal{G}}}$ of 
        $\chi_{1}^{\ideal{G}}(X)$ and $\chi_{1}^{\ideal{G}}(-X)$, respectively.
        Note that $\Lplus{\chi_{1}^{\ideal{G}}}$ and $\Lminus{\chi_{1}^{\ideal{G}}}$ consist of polynomials in $s$.
        \item Using $\Lplus{\chi_{1}^{\ideal{G}}}$, $\Lminus{\chi_{1}^{\ideal{G}}}$, make
        sequences of equations and/or inequality in $s$, such as 
        $(1,a_{d-1}>0,a_{d-2}<0,\dots,a_0>0)$. For the sequences,
        calculate the number of sign changes $\Splus{\chi_{1}^{\ideal{G}}}$ and $\Sminus{\chi_{1}^{\ideal{G}}}$
        as in \cref{eq:S+-}.
        \item By \Cref{cor:signature-cgs-qe}, collect the sequences of equations/inequalities that 
        satisfy $\Splus{\chi_{1}^{\ideal{G}}}\ne\Sminus{\chi_{1}^{\ideal{G}}}$.
        From the sequences satisfying the above condition, extract conjunction of the constraints on $s$ as $M\subset\R$.
    \end{enumerate}
\end{enumerate}

\begin{remark}
    We see that \Cref{alg:solve-ikp-cgsqe} outputs a trajectory for the given path of 
    the end-effector after verifying feasibility of the whole given path, as follows. 
    For a system of polynomial equations $F$ with parameter $s$ in 
    \cref{eq:inverse-kinematic-equations-parameters}, a CGS $\cF$ of $\ideal{F}$ is calculated.
    After calculating $\cF$, some segments without real points are eliminated optionally, 
    resulting in $\cF'$. 
    Next, for $\cF'$, the range $M$ of parameter $s$ that makes 
    \cref{eq:inverse-kinematic-equations-parameters} has a real root with the MainQE algorithm in 
    the CGS-QE method.
    Then, if $[0,1]\subset M$, a series of solution of the inverse kinematic problem $L$ is calculated by calling \Cref{alg:solve-ikp-trajectory}.
\end{remark}

We have implemented \Cref{alg:solve-ikp-cgsqe} using Risa/Asir, together with using 
Wolfram Mathematica 13.1 \cite{mathematica13} for calculating the characteristic polynomial 
in Step \ref{list:solve-ikp-cgsqe:chi}
and simplification of formula in Step \ref{list:solve-ikp-cgsqe:simplify} above.
For connecting Risa/Asir and Mathematica, OpenXM infrastructure \cite{mae-nor-oha-tak-tam2001}
was used.

\begin{example}
    Let $\p_0=\T(x_0,y_0,z_0)=\T(10,40,80)$ and $\p_f=\T(x_f,y_f,z_f)=\T(40,100,20)$
    (the same as those in \Cref{ex:time-based-path-trajectory}).
    For \cref{eq:inverse-kinematic-equations-parameters}, substitute 
    $x_0$, $y_0$, $z_0$, $x_f$, $y_f$, $z_f$ with the above values and define a system of polynomial 
    equations with parameter $s$ as 
    \begin{equation}
        \label{eq:cgs-qe-example}
        \small
        \begin{split}
            f_1 &= 120 c_1 c_4 s_7 - 16 c_1 c_4 + 120 c_1 s_4 c_7 + 136 c_1 s_4 -44 \sqrt{2} c_1 + 30 s + 10 = 0, \\
            f_2 &= 120 s_1 c_4 s_7 - 16 s_1 c_4 + 120 s_1 s_4 c_7 + 136 s_1 s_4 -44 \sqrt{2} s_1 + 60 s + 40 = 0,\\
            f_3 &= -120 c_4 s_7 - 136 c_4 + 120 s_4 s_7 -16 s_4 - 60 s - 44 \sqrt{2} - 24 = 0, \\
            f_4 &= s_1^2 + c_1^2 - 1 = 0, \quad 
            f_5 = s_4^2 + c_4^2 - 1 = 0, \quad 
            f_6 =s_7^2 + c_7^2 - 1 = 0,
        \end{split}
    \end{equation}
    and verify that \cref{eq:cgs-qe-example} has a real root for $s\in[0,1]$.
    In \Cref{alg:solve-ikp-cgsqe}, computing a CGS $\cF$ (Line \ref{line:solve-ikp-cgsqe:cgs})
    was performed in approximately $485.8$ sec., in which $\cF$ has 6 segments.
    The step of \textproc{Generate-Real-CGS} 
    (Line \ref{line:alg:solve-ikp-point:generate-real-cgs}) was performed in approximately
    $0.009344$ sec.\ with obtaining one segment existing in $\R$.
    The step of \textproc{MainQE} (Line \ref{line:solve-ikp-cgsqe:mainqe}) was
    performed in approximately $1.107$ sec., and we see that $[0,1]\subset M$, thus the
    whole trajectory is included in the feasible region of the manipulator.
    The rest of the computation is the same as the one in \Cref{ex:time-based-path-trajectory}.
\end{example}

\section{Concluding Remarks}
\label{sec:remark}
    
In this paper, we have proposed methods for inverse kinematic computation and 
path and trajectory planning of a 3-DOF manipulator using the CGS-QE method.

For the inverse kinematic computation (\Cref{alg:solve-ikp-point}), 
in addition to our previous method \cite{ota-ter-mik2021}, we have 
automated methods for eliminating segments that do not contain real points 
(\Cref{alg:Generate-Real-CGS}) and for handling non-zero dimensional 
ideals
(\Cref{sec:Solve-IKP-NonZeroDim}).
Note that our solver verifies feasibility for the given position of the end-effector before
performing the inverse kinematic computation.

For path and trajectory planning, we have proposed two methods. 
The first method (\Cref{alg:solve-ikp-trajectory}) is the repeated use of inverse kinematics solver
 (\Cref{alg:solve-ikp-point}).
The second method (\Cref{alg:solve-ikp-cgsqe}) is based on verification that the given path 
(represented as a line segment) 
is included in the feasible region of the end-effector with the CGS-QE method.
Examples have shown that the first method seems efficient and suitable for real-time solving 
of inverse kinematics problems. 
Although the second method is slower than the first one, 
it provides rigorous answers on the feasibility of path planning. 
This feature would be helpful for the initial investigation of path planning that needs rigorous 
decisions on the feasibility before performing real-time solving of inverse kinematics problems.

Further improvements of the proposed methods and future research directions include the following.
\begin{enumerate}
    \item If more than one solution of the inverse kinematic problem exist, currently we
    choose the first one that the solver returns. 
    However, currently, there is no guarantee that a series of solutions of the inverse kinematic 
    problem in the trajectory planning (in \Cref{sec:time-based-path-trajectory}) is continuous, 
    although it just so happened that the calculation in \Cref{ex:time-based-path-trajectory} 
    was well executed. 
    The problem of guaranteeing continuity of solutions to inverse kinematics problems needs to be considered in addition to the problem of guaranteeing feasibility of solutions; 
    for this purpose, tools for solving parametric semi-algebraic systems by decomposing the parametric space into connected cells above which solutions are continuous might be 
    useful \cite{che-maz2011,laz-rou2007,yan-hou-xia2001}.
    Furthermore, another criterion can be added for choosing an appropriate solution,
    based on another criteria such as the manipulability measure 
    \cite{sic-sci-vil-ori2008} that indicates how the current configuration of the joints is 
    away from a singular configuration.
    \item Our algorithm for trajectory planning (\Cref{alg:solve-ikp-trajectory}) 
    may cause a discontinuity in the sequence of the configuration of the joints when a point on the trajectory gives a non-zero dimensional ideal.
    The algorithm needs to be modified to output a sequence of continuous joint configurations, even if the given trajectory contains points that give non-zero dimensional ideals
    (see \Cref{rem:discontinuity}).
    \item Considering real-time control, the efficiency of the solver may need to be improved. 
    It would be necessary to actually run our solver on the EV3 
    to verify the accuracy and efficiency of the proposed algorithm 
    to confirm this issue (see \cref{sec:inverse-kinematics-experiments}).
    \item In this paper, we have used a line segment as a path of the end-effector. 
    Path planning using more general curves represented by polynomials would be useful 
    for giving the robot more freedom of movement.  However, if path planning becomes more complex, more efficient methods would be needed.
    \item While the proposed method in this paper is for a manipulator of 3-DOF, many
    industrial manipulators have more degrees of freedom. 
    Developing the method with our approach for manipulators of higher DOF will broaden 
    the range of applications.
\end{enumerate}

\subsubsection*{Acknowledgements}

The authors would like to thank Dr. Katsuyoshi Ohara for support for the OpenXM library to call Mathematica from Risa/Asir, and the anonymous reviewers for their helpful comments.

This research was partially supported by JSPS KAKENHI Grant Number JP20K11845.

\bibliographystyle{splncs04}
\bibliography{ev3-yoshizawa-terui-mikawa}
\end{document}